\documentclass[runningheads]{llncs}
\usepackage{graphicx}
\usepackage{cvprabbreviation}

\usepackage{tikz}
\usepackage{comment}
\usepackage{amsmath,amssymb} 
\usepackage{color}
\usepackage{graphicx}
\usepackage[ruled,vlined,noend]{algorithm2e}
\usepackage{float}
\usepackage{amsmath}
\usepackage{amssymb}
\usepackage{booktabs}
\usepackage{bbding}
\usepackage{setspace}
\usepackage{xcolor}
\usepackage{enumitem}
\usepackage{foreign}
\usepackage{subcaption}

\SetCommentSty{mycommfont}
\SetKwInput{KwInput}{Input}                
\SetKwInput{KwOutput}{Output}              
\SetKwInput{KwInit}{Initialize}
\DeclareRobustCommand*{\ora}{\overrightarrow}
\DeclareRobustCommand*{\ola}{\overleftarrow}

\usepackage[pagebackref=true,breaklinks=true,letterpaper=true,colorlinks,bookmarks=false]{hyperref}

\usepackage[accsupp]{axessibility}  

\usepackage{cite}

\begin{document}
\pagestyle{headings}
\mainmatter
\def\ECCVSubNumber{4024}  

\title{Unidirectional Video Denoising by Mimicking Backward Recurrent Modules with Look-ahead Forward Ones} 

\titlerunning{Unidirectional Video Denoising}
%
\author{Junyi Li\inst{1} \and
Xiaohe Wu\inst{1} \and
Zhenxing Niu\inst{2} \and Wangmeng Zuo\inst{1}}
\authorrunning{J. Li et al.}
%
\institute{Harbin Institute of Technology \and
Xidian University\\
\email{\{nagejacob, csxhwu, zhenxingniu\}@gmail.com}\\ 
\email{wmzuo@hit.edu.cn}
}
\maketitle
\begin{center}
\centering
\captionsetup{type=figure}
\includegraphics[width=\linewidth]{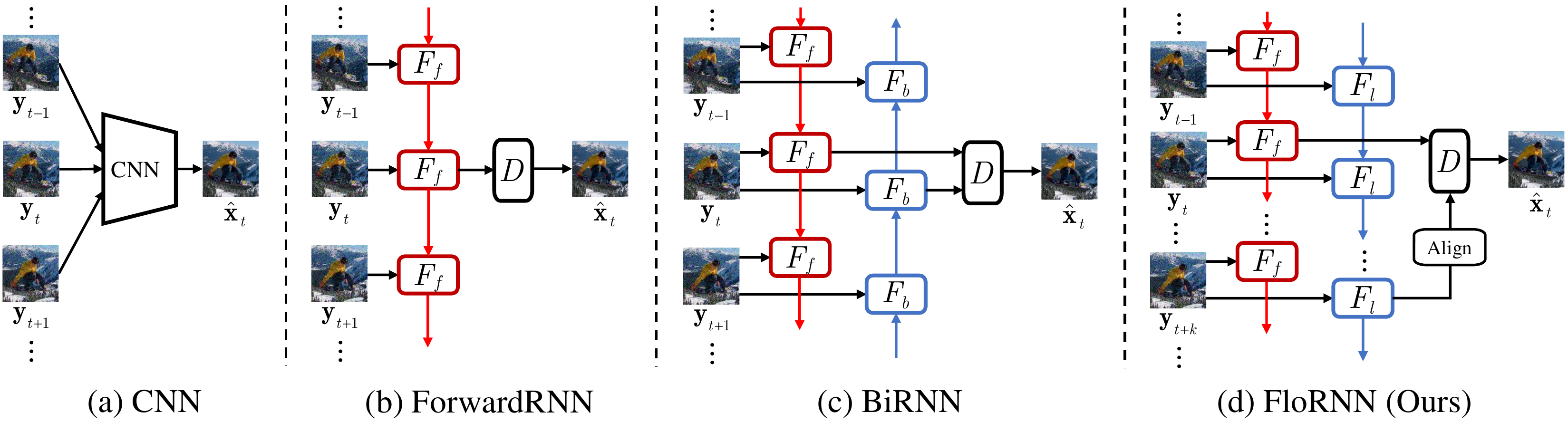}
\caption{Illustration of representative deep network architectures for video denoising. 
When handling the current frame, (a) CNN cannot exploit long-term temporal information, and (b) ForwardRNN cannot exploit any future frames. 
(c) BiRNN is effective in exploiting the information from all frames, but can only be performed in an \emph{offline} manner.  
In comparison, (d) FloRNN can leverage both the historical information and the crucial near-future frames, thereby being very appealing for \emph{unidirectional} video denoising.}
\label{fig:overview}
\end{center}

\begin{abstract}
While significant progress has been made in deep video denoising, it remains very challenging for exploiting historical and future frames.
Bidirectional recurrent networks (BiRNN) have exhibited appealing performance in several video restoration tasks.
However, BiRNN is intrinsically offline because it uses backward recurrent modules to propagate from the last to current frames, which causes high latency and large memory consumption.
To address the offline issue of BiRNN, we present a novel recurrent network consisting of forward and look-ahead recurrent modules for unidirectional video denoising.
Particularly, look-ahead module is an elaborate forward module for leveraging information from near-future frames.
When denoising the current frame, the hidden features by forward and look-ahead recurrent modules are combined, thereby making it feasible to exploit both historical and near-future frames.
Due to the scene motion between non-neighboring frames, border pixels missing may occur when warping look-ahead feature from near-future frame to current frame, which can be largely alleviated by incorporating forward warping and proposed border enlargement.
Experiments show that our method achieves state-of-the-art performance with constant latency and memory consumption.
Code is avaliable at \url{https://github.com/nagejacob/FloRNN}.

\end{abstract}

\section{Introduction}
Recent years have witnessed the great success of deep networks in video denoising~\cite{claus2019videnn,tassano2019dvdnet,tassano2020fastdvdnet,davy2019non,davy2021video,vaksman2021patch,yue2020supervised,maggioni2021efficient,xu2020learning}. 
Compare to image denoising, temporal information plays a pivotal role in video denoising, which is generally restricted by the spatial misalignment among consecutive frames.
Self-similar spatial-temporal patch aggregation has been suggested for spatio-temporal modeling\cite{davy2021video,vaksman2021patch}, but usually results in heavy computational cost. 
Other spatio-temporal alignment methods, \eg, optical flow~\cite{sun2018pwc}, deformable convolution~\cite{dai2017deformable}, and kernel prediction network (KPN)~\cite{mildenhall2018burst}, have also been studied and applied for video denoising~\cite{tassano2019dvdnet,yue2020supervised,mildenhall2018burst,xia2020basis,xu2020learning}. 
Instead of explicit motion estimation and compensation, cascaded U-Net~\cite{tassano2020fastdvdnet} and multi-stage recurrent network~\cite{maggioni2021efficient} are further suggested for the efficiency issue. 
Albeit the progress in deep video denoising, it remains a challenging issue to exploit historical and future frames for video denoising.

Bidirectional recurrent networks (BiRNN) provide a convenient way for temporal modeling and have been very appealing in several video restoration tasks
\cite{chan2021basicvsr,chan2021basicvsr++,son2021recurrent,kim2019recurrent,shen2020blurry,yang2021recurrent}. 
In video super-resolution (VSR), BasicVSR~\cite{chan2021basicvsr} and its extension~\cite{chan2021basicvsr++} have outperformed most state-of-the-art methods in terms of VSR performance and efficiency. 
Benefited from the forward and backward recurrent modules, BiRNN is effective in exploiting the information from all frames to restore the current frame.
In comparison, convolutional network (CNN) takes neighborhood frames as inputs and only exploits short-term temporal information. 
However, BiRNN is intrinsically an offline approach where the backward recurrent module is deployed to propagate information from the last frame to current frame, and the restoration result of current frame cannot be obtained unless all video frames are processed.
Meanwhile, hidden features of all frames have to be maintained in the memory during inference, which causes high memory consumption.
The high latency and large memory consumption
limit the practicability of BiRNN.

To address the offline issue of BiRNN, we present a novel recurrent network consisting of \textbf{F}orward and \textbf{lo}ok-ahead recurrent modules (\ie, FloRNN) for unidirectional video denoising.
From~\cite{chan2021basicvsr}, the future frames are important to boost the denoising performance of current frame.
Simply discarding backward modules from BiRNN (denote as ForwardRNN) hampers the use of future frames and results in inferior performance.
Moreover, among all future frames, the near-future frames are most important to enhance the denoising results of current frame.
%
%
To leverage future information while addressing offline issue, we present a look-ahead recurrent module for exploiting near-future frames.
%
%
As shown in Fig.~\ref{fig:overview}(d), our look-ahead module $F_l$ adopts a forward recurrent architecture, but propagates $k$ frames ahead of the forward module $F_f$. 
The look-ahead feature of the near-future frame (\ie, $t+k$) is warped back to align with current frame (\ie, $t$), and incorporated with the forward feature from $F_f$ to produce current denoising results.
Our FloRNN adopts a unidirectional propagation, yet extracts crucial near-future frames with $F_l$.
Thus, it has the potential to approach the performance of BiRNN while maintaining constant latency and memory consumption.
Furthermore, one can adjust $k$ to meet the latency constraint, smaller $k$ for real-time tasks, and larger $k$ for better results.

Nonetheless, there remains a major issue to address for applying look-ahead module.
As shown in Fig.~\ref{fig:border}(a), with an assumption of $k$ = 2, the motion between the $t$-th and $(t+2)$-th frames makes that an amount of border pixels are lost when warping the look-ahead feature from $(t+2)$-th frame back to $t$-th frame.
Obviously, such insufficient utilization of near-future frames is harmful to denoising performance.
%
%
To address this issue, we present to incorporate forward warping and border enlargement mechanism.
As shown in Fig.~\ref{fig:border}(b), forward warping is adopted in the look-ahead module instead of backward warping.
Meanwhile, we suggest to enlarge the border of look-ahead feature to a fixed ratio (\eg, 10\% in this work) during the forward warping.
Due to that forward warping allows warped feature to splat out of border (\ie, pixels out of the red box), we can save the missed border pixels when warping look-ahead feature to future frames, and recover them when aligning look-ahead feature back to the current frame.
Our forward warping and border enlargement mechanism can largely reduce the amount of missed border pixels for better exploiting near-future frames and also benefiting denoising performance.

Extensive experiments are conducted to evaluate our FloRNN on commonly used video denoising benchmarks.
Both synthetic additive white Gaussian noise (AWGN) and real-world noise are considered in our experiments.
For AWGN, our method outperforms the second best competing method PaCNet~\cite{vaksman2021patch} by a large margin (i.g., 0.76dB by PSNR) on DAVIS. 
On real-world video datasets such as CRVD~\cite{yue2020supervised} and IOCV~\cite{kong2020comprehensive}, FloRNN also achieves the best quantitative results than the competing methods.
We also show that our look-ahead module could be applied to state-of-the-art BiRNN methods (\eg, BasicVSR++~\cite{chan2021basicvsr++}), which forms a unidirectional counterpart and benefits from the advances in BiRNNs in the supplementary material.

To sum up, the main contribution of this work includes:
\begin{itemize}

    \item A novel recurrent network (\ie, FloRNN) is presented for unidirectional video denoising by incorporating forward and look-ahead recurrent modules. 
    
    \item Forward warping and border enlargement are equipped in look-ahead recurrent module for better exploiting the near-future frames and also benefiting denoising performance.
    
    \item Experiments show that our method performs favorably against state-of-the-arts on various video denoising datasets.
\end{itemize}

\section{Related Work}
    
\subsection{Image Denoising.} With the observation that an image patch usually has many similar counterparts within the same image, many traditional methods  are developed for joint modeling of a stack of similar patches to remove noise, such as non-local means (NLM)~\cite{buades2005non}, BM3D~\cite{dabov2007image}, and WNNM~\cite{gu2014weighted}. 
Other methods use handcrafted priors on image patches, \eg, sparsity~\cite{elad2006image} and Gaussian Mixture~\cite{zoran2011from}. 
Recently, CNN based methods have achieved favorable performance.
DnCNN~\cite{zhang2017beyond} incorporates residual learning~\cite{he2016deep} and batch normalization~\cite{ioffe2015batch} for better convergence.
FFDNet~\cite{zhang2018ffdnet} takes noise level map as input, and trains a single model for handling various noise levels.
With the collected real-world denoising datasets~\cite{plotz2017benchmarking,abdelhamed2018high}, several methods~\cite{guo2019toward,zhuo2019ridnet,kim2020transfer} have also been suggested to handle real noise.

\subsection{Video Denoising.} 
For exploiting temporal information, traditional patch-based methods~\cite{maggioni2012video,arias2018video,buades2016patch} search similar patches on volumetric data, and are very time consuming. 
By leveraging deep image denoising, several methods adopt two-stage scheme including spatial denoising and temporal fusion.
ViDeNN~\cite{claus2019videnn} uses a plain CNN as fusion network, while DVDNet~\cite{tassano2019dvdnet} assists the fusion with optical flow and warping neighbor frames to current frame. 
Temporal aggregation mechanisms have also been investigated and  integrated into network design, \eg, non-local search~\cite{davy2021video,vaksman2021patch}, cascaded U-Net~\cite{tassano2020fastdvdnet}, kernel prediction network (KPN)~\cite{mildenhall2018burst,xia2020basis,xu2020learning}, deformable convolution~\cite{yue2020supervised}, and channel shifting~\cite{rong2020burst}. 
However, these aggregation mechanisms either are complicated and inefficient, or cannot achieve state-of-the-art performance.
EMVD~\cite{maggioni2021efficient} proposes a multi-stage recurrent architecture for mobile devices, but performs inferior when scaling to larger model on GPUs. 
    
\subsection{RNN in Video Restoration.} 
RNN provides a convenient way for temporal modeling and thus can be readily applied to video restoration~\cite{chen2016deep,huang2017video}. 
Taking VSR as an example, FRVSR~\cite{sajjadi2018frame} warps the $(t-1)$-th output to $t$-th frame, and use it as additional input to super-resolve the $t$-th frame.
RLSP~\cite{fuoli2019efficient} introduces high-dimentional latent states for efficient propagation.
RSDN~\cite{isobe2020video} divides the input into structure and detail components to effectively exploit temporal correlation.
BasicVSR~\cite{chan2021basicvsr} adopts bidirectional propagation and optical flow based alignment, achieving state-of-the-art VSR performance.
BasicVSR++\cite{chan2021basicvsr++} further improves BasicVSR with flow-guided deformable alignment and second order propagation. 
However, bidirectional propagation in BasicVSR~\cite{chan2021basicvsr} and its extension\cite{chan2021basicvsr++} makes them unable to be used for online video restoration.
RNN has also been investigated for video deblurring~\cite{nah2019recurrent,son2021recurrent,zhong2020efficient}, video inpainting~\cite{kim2019deep,kim2019recurrent}, video frame interpolation~\cite{shen2020blurry,xiang2020zooming}, and video deraining~\cite{liu2018erase,yang2019frame,yang2021recurrent}.

\begin{figure}[t]
\centering
\includegraphics[width=\linewidth]{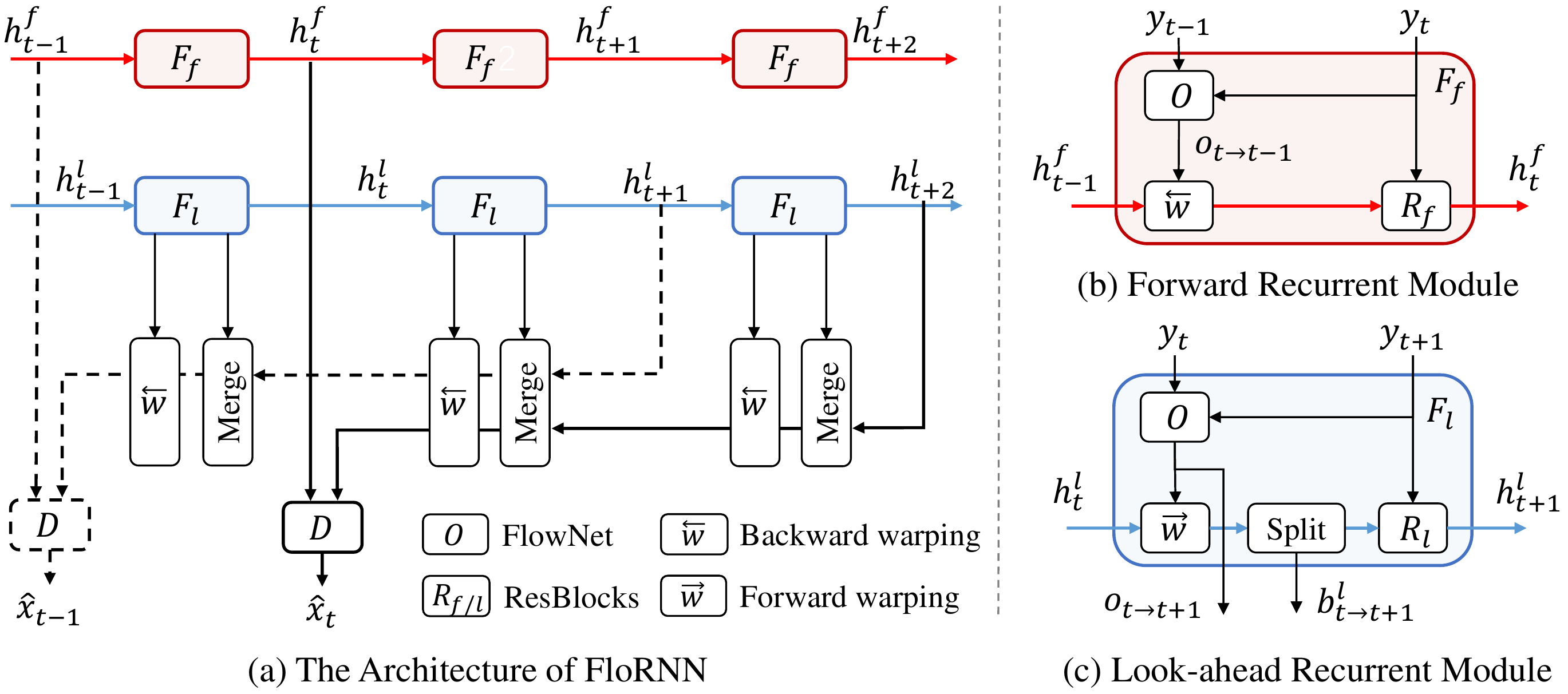}
\caption{\textbf{Illustration of FloRNN}. (a) FloRNN is mainly built with three components, \ie, forward recurrent module $F_f$, look-ahead recurrent module $F_l$ and decoder $D$. (b) $F_f$ propagates from first to current frames which exploits all previous frames, (c) $F_l$ propagates similarly to $F_f$ in a forward manner, but $k$ (\eg, $k$ = 2) frames ahead of $F_f$. The look-ahead feature is aligned back to current frame (details of Split and Merge operations are provided in Fig.~\ref{fig:border} (b)) for exploiting crucial near-future frames. With the propagated historical and near-future information, FloRNN achieves compelling results while has constant latency and memory occupation.}
\label{fig:method}
\end{figure}

\section{Method}
Given a video sequence consisting of $T$ noisy frames $\{\mathbf{y}_t\}_{t=1}^T$, video denoising aims to produce the prediction $\{\hat{\mathbf{x}}_t\}_{t=1}^T$ for approximating its clean video $\{\mathbf{x}_t\}_{t=1}^T$.
For better denoising of current frame $\mathbf{y}_t$, one favorable solution is to exploit all frames $\{\mathbf{y}_t\}_{t=1}^T$ to predict $\hat{\mathbf{x}}_t$. 
BasicVSR~\cite{chan2021basicvsr} originally suggested for VSR provides a bidirectional recurrent network (BiRNN, as shown in Fig.~\ref{fig:overview}(c) ) to leverage the information from all frames, and we empirically find that it is also very appealing for video denoising.
However, due to the use of backward recurrent module, all the succeeding frames are required to produce backward hidden feature, making that BiRNN can only be performed in an offline manner.

To address the \emph{offline} issue of BiRNN, we propose a novel recurrent network for \emph{unidirectional} video denoising.
As shown in Fig.~\ref{fig:overview}(d), it consists of a \textbf{F}orward recurrent module as well as a \textbf{lo}ok-ahead recurrent module, named as FloRNN.
Analogous to BiRNN, FloRNN adopts the same forward recurrent module $F_f$ for history frames propagation and decoder $D$ for producing denoising results.
The difference is that FloRNN substitutes the backward recurrent module of BiRNN with an elaborate look-ahead recurrent module $F_l$ for exploiting near-feature frames.
%
%
The look-ahead feature from $F_l$ is then aligned and incorporated with forward feature from $F_f$ to enhance the denoising results of current frame. 
To address the border missing issue during the alignment, we further introduce a border enlargement mechanism with forward warping.
The details of our framework are illustrated in Fig.~\ref{fig:method}.
In the following, we will describe the forward module, look-ahead module, forward warping and border enlargement, respectively.

\subsection{Forward Recurrent Module}
%
Analogous to BiRNN, we first introduce the forward recurrent module $F_f$ to propagate information from the first to the current frames.
As shown in Fig.~\ref{fig:method}(b), it adopts a recurrent manner by propagating the forward hidden feature $\mathbf{h}_{t-1}^f$  and combining it with the current frame $\mathbf{y}_t$ to obtain the hidden feature $\mathbf{h}_{t}^f$ at frame $t$.
\begin{equation}
\label{equ:Ff}
    \mathbf{h}^f_t = F_f(\mathbf{y}_t, \mathbf{y}_{t-1}, \mathbf{h}^f_{t-1}).
\end{equation}
Nonetheless, $\mathbf{h}_{t-1}^f$ is aligned with $\mathbf{y}_{t-1}$ instead of $\mathbf{y}_{t}$.
Following~\cite{chan2021basicvsr}, we estimate the optical flow~\cite{sun2018pwc} from frame $t$ to $t-1$, which is used to align and aggregate $\mathbf{h}_{t-1}^f$ with $\mathbf{y}_{t}$. 
Thus, the forward recurrent module in Eqn.~(\ref{equ:Ff}) can be further written as,
\begin{equation}
\label{eqn:forRec}
\begin{split}
    \mathbf{o}_{t\rightarrow t-1} &= O(\mathbf{y}_t, \mathbf{y}_{t-1}),\\
    \mathbf{h}^{f}_{t-1\rightarrow t} &= \overleftarrow{w}(\mathbf{h}^{f}_{t-1}, \mathbf{o}_{t\rightarrow t-1}),\\
    \mathbf{h}^{f}_t &= R_{f}(\mathbf{y}_t, \mathbf{h}^{f}_{t-1\rightarrow t}).
\end{split}
\end{equation}
where $O(\cdot, \cdot)$ denotes an optical flow network, and $\mathbf{o}_{t\rightarrow t-1}$ denotes the estimated optical flow from $t$-th frame to ($t$-1)-th frame.
$\overleftarrow{w}(\mathbf{h}^{f}_{t-1}, \mathbf{o}_{t\rightarrow t-1})$ stands for aligning $\mathbf{h}^{f}_{t-1}$ with $\mathbf{y}_t$ using backward warping~\cite{chan2021basicvsr} to obtain the warped hidden feature $\mathbf{h}^{f}_{t-1\rightarrow t}$.
$R_{f}$ aggregates the warped hidden feature and current frame with multiple residual blocks to obtain the hidden feature $\mathbf{h}^{f}_t$ of $t$-th frame.

\subsection{Look-ahead Recurrent Module}
\label{Look-ahead}
%
%
To leverage the future information while addressing the offline issue, we propose the look-ahead recurrent module for exploiting near-future frames.
As shown in Fig.~\ref{fig:overview}(d), our look-ahead module adopts a specifically designed forward recurrent mechanism,
\begin{equation}
\label{lookaheadf}
\mathcal{H}^l_{t+1} = F_l(\mathbf{y}_{t+1}, \mathbf{y}_{t}, \mathcal{H}^l_{t}).
\end{equation}
Here we use $\mathcal{H}^l_{t+1}$ to indicate that the output of look-ahead module may contain hidden feature $\mathbf{h}^l_{t+1}$ and other variables.
As the look-ahead module propagates $k$ frames ahead of the forward module, $\mathbf{h}^l_{t+k}$ captures temporal information of $k$ near-future frames. 
For restoring $t$-th frame,
we align the look-ahead feature $\mathbf{h}^l_{t+k}$ back to $t$-th frame to produce the warped look-ahead feature $\mathbf{h}^l_{t+k\rightarrow t}$, which can be generally written as,
\begin{equation}
\begin{aligned}
    \mathbf{h}^l_{t+k\rightarrow t} = 
    \text{Align}\left(\mathbf{h}^l_{t+k},\ \cdots\right).
\end{aligned}
\end{equation}
With the forward feature $\mathbf{h}^f_t$ and the aligned look-ahead feature $\mathbf{h}^l_{t+k\rightarrow t}$, the denoising result at frame $t$ can then be obtained by,
\begin{equation}
\hat{\mathbf{x}}_t = D(\mathbf{h}^f_t, \mathbf{h}^l_{t+k\rightarrow t}).
\end{equation}
where the decoder $D$ contains two convolution layers.

Note that $F_l$ and $\text{Align}(\mathbf{\cdot},\ \cdots)$ can be implemented with different forms. 
For example, one can adopt the forward recurrent module defined in Eqn.~(\ref{eqn:forRec}) to implement $F_l$. 
As for $\text{Align}(\mathbf{\cdot},\ \cdots)$, one direct solution is to compute the optical flow from $\mathbf{y}_{t+k}$ to $\mathbf{y}_t$, then warp $\mathbf{h}^l_{t+k}$ back to $\mathbf{y}_t$. 
In the following subsection, forward warping and border enlargement are presented as a reasonable implementation.

\begin{figure}[t]
\centering
\includegraphics[width=\linewidth]{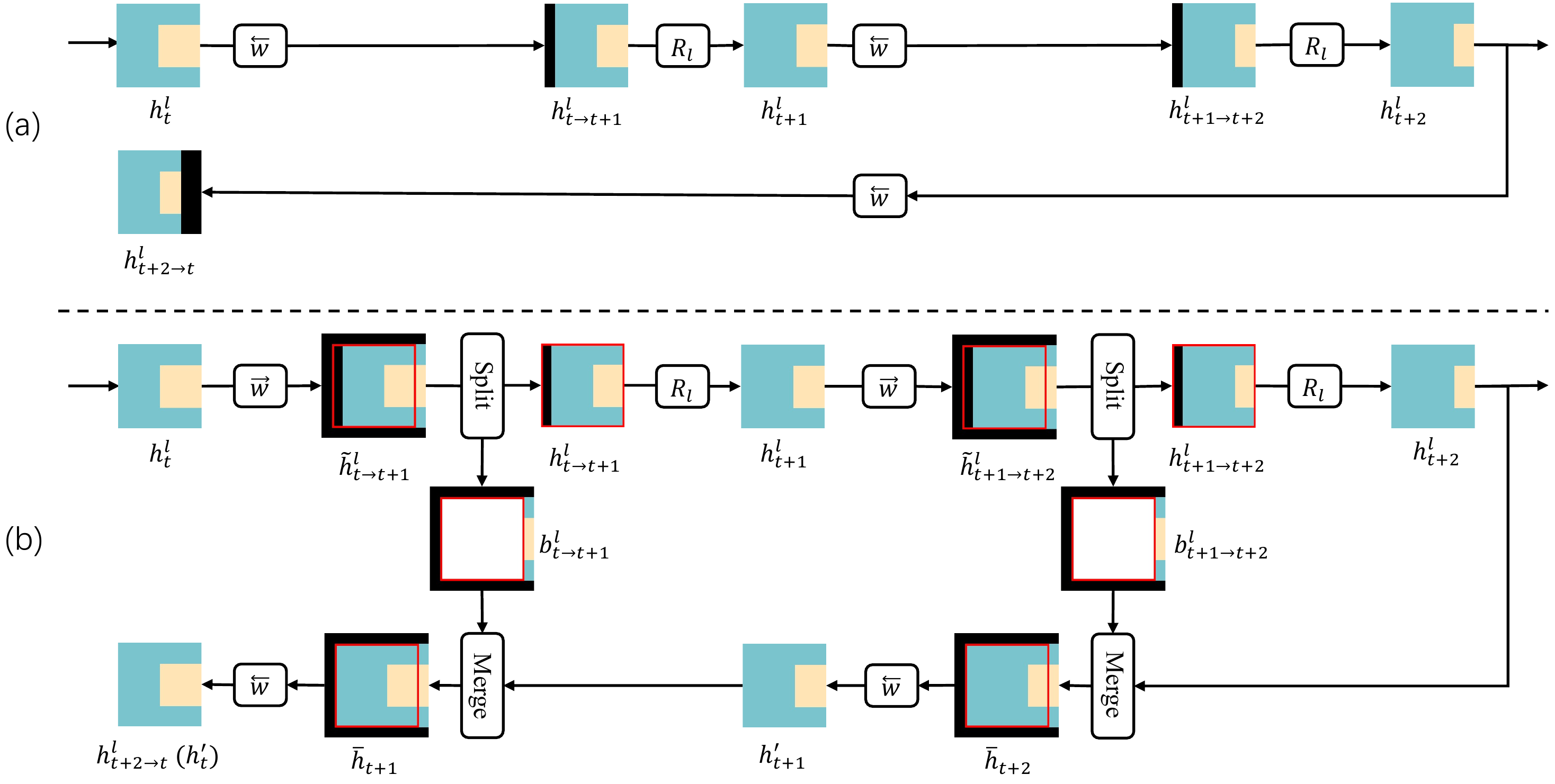}
\caption{Illustration of forward warping and border enlargement for our look-ahead recurrent module. (a) Aligning $h^l_{t+2}$ to $t$-th frame with backward warping $\ola{w}$ suffers from border pixels missing problem. (b) Forward warping $\ora{w}$ allows the out of border pixels to splat to the enlarged border. We save the enlarged border when warping to future frames and recover it when warping back, which largely mitigates the border information missing problem.}
\label{fig:border}
\end{figure}
\subsection{Forward Warping and Border Enlargement}
\begin{figure}[t]
\centering
\includegraphics[width=\linewidth]{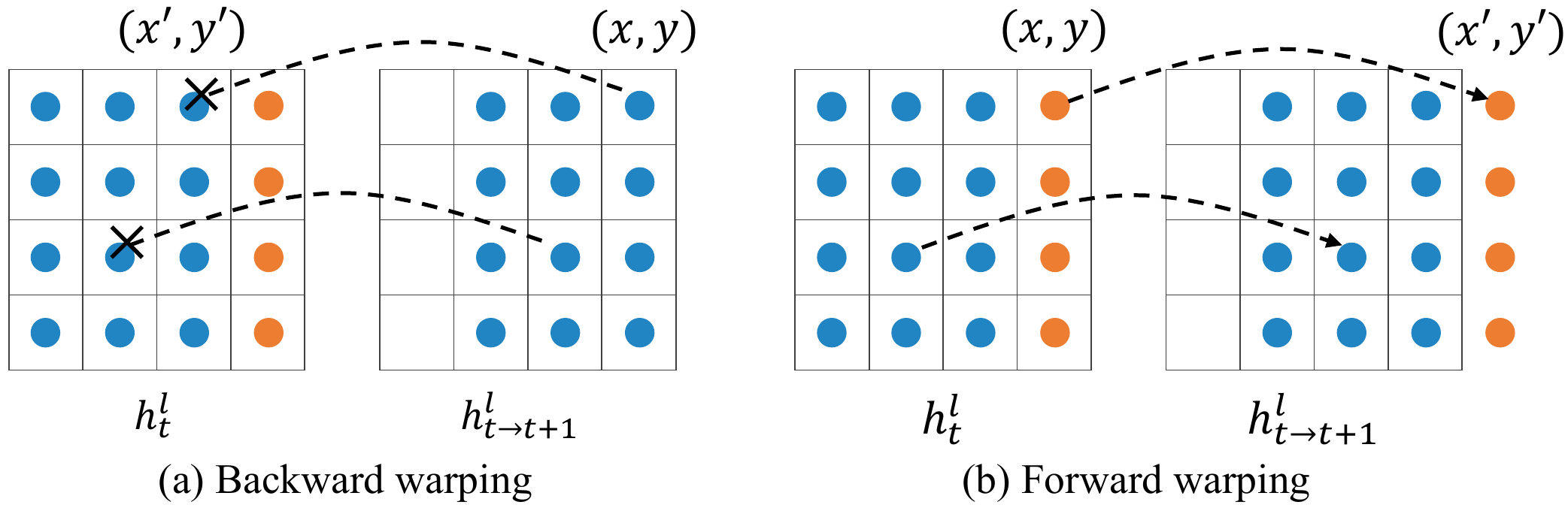}
\caption{Border pixels (colored in orange) in $\mathbf{h}^l_t$ are out of border in the $(t+1)$-th frame. (a) Backward warping ignores these border pixels in the warping result $\mathbf{h}^l_{t\to t+1}$, (b) forward warping splats these pixels to the outside of border, which can be preserved with border enlargement mechanism.}
\label{fig:warping}
\end{figure}

As stated above, one straightforward way to implement the look-ahead module is adopting the forward recurrent module defined in Eqn.~(\ref{eqn:forRec}) as $F_l$ and aligning look-ahead feature to current frame by backward warping. 
However, as shown in Fig.~\ref{fig:border}(a), such a straightforward implementation suffers from border information missing issue when aligning look-ahead feature back to current frame (\ie, black part exists in $\mathbf{h}^l_{t+2 \to t}$). 
This is caused by scene motion between consecutive frames and backward warping mechanism in look-ahead module.
To demonstrate this, we first give a brief introduction of backward warping.
As shown in Fig.~\ref{fig:warping}(a), taking warping $\mathbf{h}^l_{t}$ to $\mathbf{h}^l_{t\rightarrow t+1}$ as example, the value of $\mathbf{h}^l_{t\rightarrow t+1}$ at position $(x,y)$ is sampled from $\mathbf{h}^l_{t}$ at position $(x^{\prime}, y^{\prime})$ according to backward optical flow $(x^{\prime}, y^{\prime}) = \left(\mathbf{o}_{t+1\rightarrow t}\right)_{x,y}$.
When scene motion exists, border pixels in $\mathbf{h}^l_{t}$ may have no correspondence in $\mathbf{h}^l_{t+1}$, and are implicitly dropped in the warped result.
Along with the increase of timestamps, the part of lost information of current frame in look-ahead feature becomes larger.
Consequently, when aligning $\mathbf{h}^l_{t+k}$ to current frame $t$, the lost information cannot be recovered (as shown in Fig.~\ref{fig:border}(a)). 
Such information missing leads to unexpected inferior performance.

To address the above issue, we replace backward warping in the look-ahead module with forward warping incorporated with border enlargement mechanism.
As shown in Fig.~\ref{fig:warping}(b), forward warping splats the value of $\mathbf{h}^{l}_{t}$ at position $(x,y)$ to $\mathbf{h}^{l}_{t\rightarrow t+1}$ at position $(x^{\prime}, y^{\prime})$ according to forward optical flow $(x^{\prime}, y^{\prime}) = \left(\mathbf{o}_{t\rightarrow t+1}\right)_{x,y} $.
When scene motion exists, border pixels in $\mathbf{h}^l_{t}$ are splatted to positions out of the border.
This offers an opportunity to save and reuse the out of border pixels by border enlargement mechanism as illustrated in Fig.~\ref{fig:border}(b).
Specifically, forward flow is first calculated for aligning the $\mathbf{h}^l_t$ to ($t$+1)-th frame,
\begin{equation}
\begin{aligned}
    \mathbf{o}_{t\rightarrow t+1} &= O(\mathbf{y}_{t}, \mathbf{y}_{t+1}).
\end{aligned}
\end{equation}
To address the border missing issue, we first enlarge the border of the feature to a certain percent (\eg, 10\%) and then perform forward warping,
\begin{equation}
\begin{aligned}
    \tilde{\mathbf{h}}^l_{t \rightarrow t+1} &= \overrightarrow{w}(\mathbf{h}^l_{t}, \mathbf{o}_{t\rightarrow t+1}).
\end{aligned}
\end{equation}
The warped feature $\tilde{\mathbf{h}}^{l}_{t \rightarrow t+1}$ is further split into two separate parts,
\begin{equation}
    \{\mathbf{b}^l_{t\rightarrow t+1}, \mathbf{h}^l_{t\rightarrow t+1}\} =  \textit{Split}(\tilde{\mathbf{h}}^l_{t\rightarrow t+1}).
\end{equation}
$\mathbf{b}^l_{t\rightarrow t+1}$ denotes the enlarged border region that contains the pixels splatting out of border. $\mathbf{h}^l_{t\rightarrow t+1}$ denotes the within-frame region aligned with $(t+1)$-th frame, and is further aggregated with $\mathbf{y}_{t+1}$ to produce the hidden feature $\mathbf{h}_{t+1}^{l}$ with multiple residual blocks $R_l$,
\begin{equation}
    \mathbf{h}^l_{t+1}  = R_l(\mathbf{y}_{t+1}, \mathbf{h}^l_{t\rightarrow t+1}).
\end{equation}
%
%
In this manner, border pixels are preserved in the enlarged border $\mathbf{b}^l_{t\rightarrow t+1}$, which could be reused to recover the missed border region when aligning the look-ahead feature $\mathbf{h}^l_{t+k}$ to current frame.
In particular, $\mathbf{h}^l_{t+k}$ is warped to $t$-th frame in a frame by frame manner.
We first initialize $\mathbf{h}^{\prime}_{t+k} = \mathbf{h}^{l}_{t+k}$.
When warping from ($t$+$i$)-th frame to ($t$+$i$-1)-th frame, we first merge the border information $\mathbf{b}^l_{t+i-1\rightarrow t+i}$ with $\mathbf{h}^{\prime}_{t+i}$.
\begin{equation}
    \bar{\mathbf{h}}_{t+i} = \textit{Merge}(\mathbf{h}^{\prime}_{t+i}, \mathbf{b}^l_{t+i-1\rightarrow t+i})
\end{equation}
Then, $\bar{\mathbf{h}}_{t+i}$ is aligned to ($t$+$i$-1)-th frame with backward warping,
\begin{equation}
\mathbf{h}^{\prime}_{t+i-1} = \overleftarrow{w}(\bar{\mathbf{h}}_{t+i}, \mathbf{o}_{t+i-1\rightarrow t+i}).
\end{equation}
After $k$ steps, the warped look-ahead feature can be obtained as,
\begin{equation}
\begin{aligned}
    \mathbf{h}^l_{t+k\rightarrow t} = \mathbf{h}^{\prime}_{t}.
\end{aligned}
\end{equation}
From Fig.~\ref{fig:border}(b), with the proposed border enlargement mechanism, the aligned look-ahead feature $h^l_{t+2 \to t}$ recovers the missed border information, which brings obvious performance gain. 
Moreover, we note that the optical flow $\mathbf{o}_{t+i-1\rightarrow t+i}$ reuses the optical flow for forward warping in look-ahead module (\ie, $\mathcal{H}^l_{t+1} = \{ \mathbf{h}^l_{t+1}, \mathbf{b}_{t\rightarrow t+1}, \mathbf{o}_{t\rightarrow t+1}\}$), thereby being effective in improving consistency and efficiency in aligning look-ahead feature back.

\section{Experiments}
\label{sec:exp}
\begin{figure}[t]
    \centering
\begin{subfigure}[b]{0.45\textwidth}
    \centering
    \includegraphics[width=\textwidth]{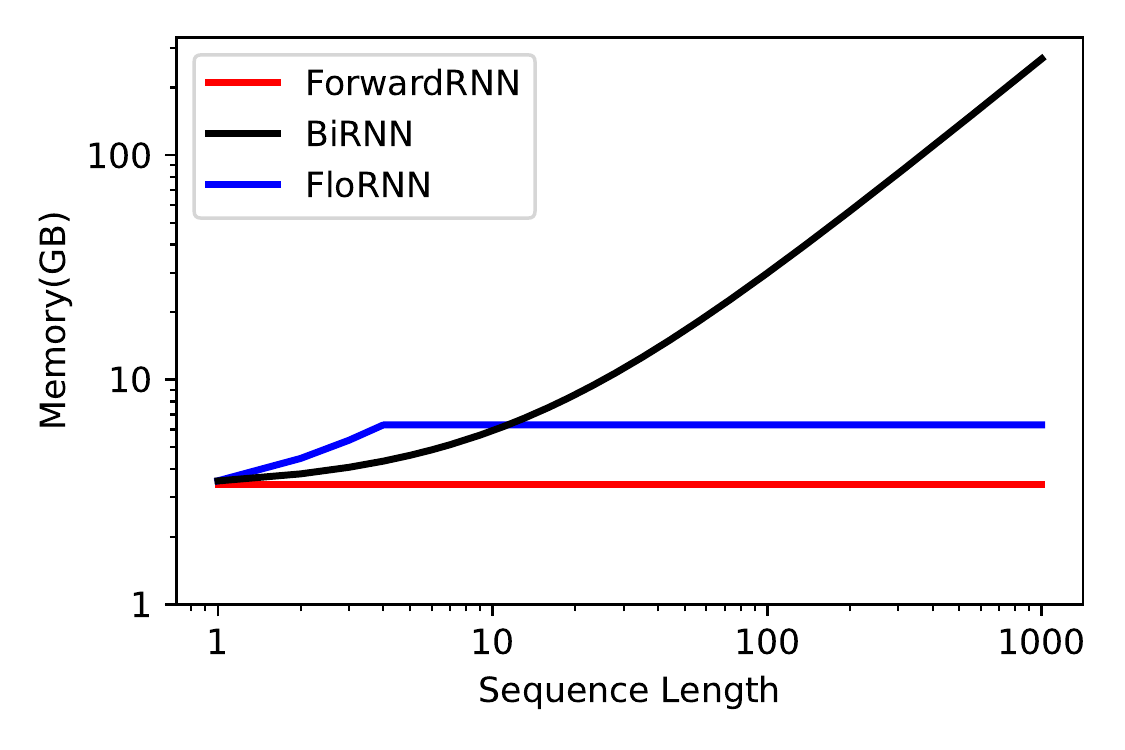}
\end{subfigure}
\hfill
\begin{subfigure}[b]{0.45\textwidth}
    \centering
    \includegraphics[width=\textwidth]{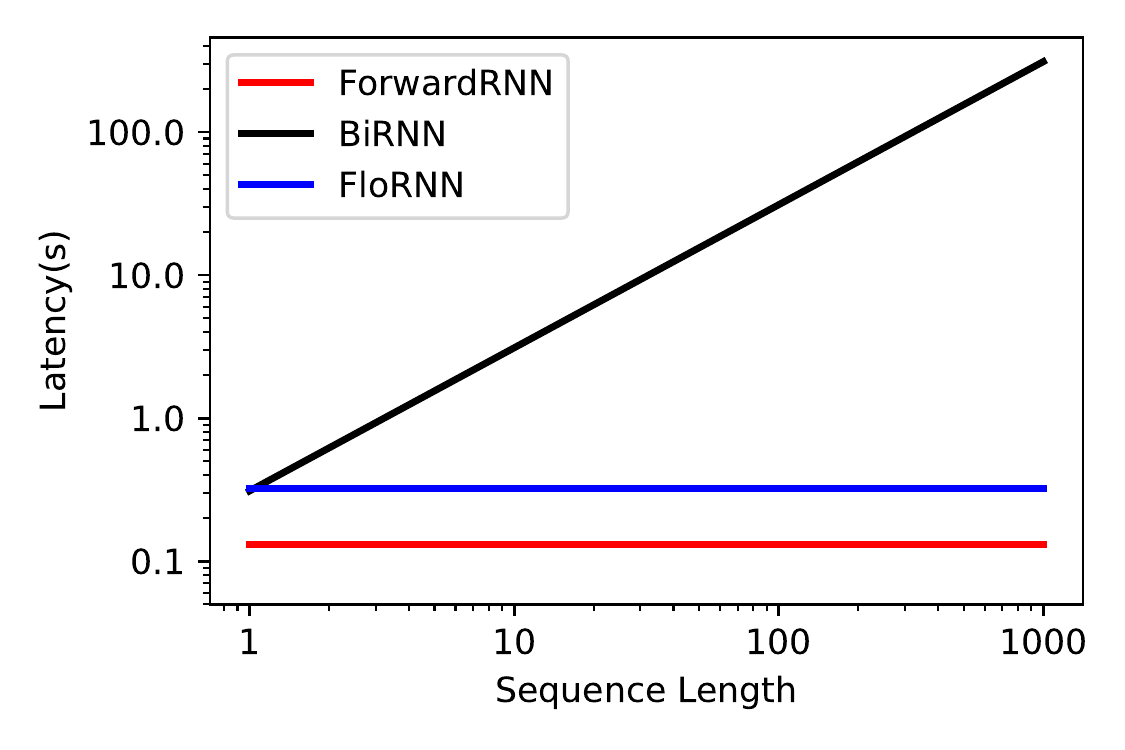}
\end{subfigure}
\caption{Analysis of memory consumption (left) and latency (right) for three representative recurrent methods.}
\label{fig:memory}
\end{figure}
\begin{table}[t]
\scriptsize
\caption{Quantitative comparison of three representative recurrent methods on Set8~\cite{tassano2019dvdnet}, FloRNN performs close to BiRNN while maintaining unidirectional.}
\centering
\setlength{\tabcolsep}{3.6mm}{
\begin{tabular}{@{}cccccc@{}}
\toprule
Model & Unidirectional & Online & PSNR/SSIM\\
\midrule
ForwardRNN & \CheckmarkBold & \CheckmarkBold & 33.12/0.9089\\
FloRNN & \CheckmarkBold & \XSolidBrush & 33.55/0.9153\\
BiRNN & \XSolidBrush & \XSolidBrush & 33.74/0.9192\\
\bottomrule
\end{tabular}}
\label{tab:ablation_architecture}
\end{table}

\subsection{Experimental Settings} 
\noindent \textbf{Datasets.}  To evaluate our method on both synthetic and real-world noisy videos, we conduct experiments on the following datasets,

\begin{itemize}[leftmargin=*]
    \setlength{\itemsep}{0pt}
	\setlength{\parsep}{0pt}
	\setlength{\parskip}{0pt}
    \item Set8~\cite{tassano2019dvdnet} and DAVIS~\cite{pont20172017} are two widely used synthetic Gaussian video denoising datasets.
    \item CRVD~\cite{yue2020supervised} is a real-world video denoising dataset captured in raw domain. It contains 6 indoor scenes for training, 5 indoor scenes for testing, and 10 dynamic outdoor scenes without ground-truth for visual evaluation.
    \item IOCV~\cite{kong2020comprehensive} is a real-world video denoising test set in sRGB domain. Each noisy video is captured by a smartphone multiple times, and the mean video is taken as ground-truth.
\end{itemize}

\begin{table}[t]
\caption{Ablation study for look-ahead recurrent module on Set8 dataset~\cite{tassano2019dvdnet}.}
\scriptsize
\begin{subtable}[c]{\linewidth}
    \centering
    \caption*{(a) Effects on different alignment mechanism.}
    \setlength{\tabcolsep}{3.4mm}
    \begin{tabular}{ccc}
    \toprule
    Warping & Border Enlargement & PSNR/SSIM\\
    \midrule
    Backward warping & \XSolidBrush & 33.44/.9132\\
    Forward warping & \XSolidBrush & 33.45/.9134\\
    Forward warping & \CheckmarkBold & 33.55/.9153\\
    \bottomrule
    \end{tabular}
\end{subtable}
\begin{subtable}[c]{\linewidth}
    \centering
    \caption*{(b) Effects on the number of near-future frames $k$.}
    \setlength{\tabcolsep}{2.7mm}
    \begin{tabular}{ccccccc}
    \toprule
    $k$ & 0 & 1 & 2 & 3 & 4 & 5\\
    \midrule
    PSNR & 33.12 & 33.47 & 33.53 & 33.55 & 33.53 & 33.51\\
    \bottomrule
    \end{tabular}
\end{subtable}
\label{tab:ablation_components}
\end{table}

\noindent \textbf{Implementation Details.} 
We adopt a pretrained PWC-Net~\cite{sun2018pwc} as our optical flow network and fix the parameters of flow network.
We find that PWC-Net generalizes well on noisy data, perhaps benefiting from additive Gaussian noise data augmentation during training~\cite{dosovitskiy2015flownet}.
For raw videos, we use the demosaiced frames as inputs to PWC-Net.
Training sequences are cropped at random spatial-temporal locations, with spatial patch size 96$\times$96.
The batch size is set to 16 and training length $T$ = $10$.
We use $\ell_2$ loss to train our network, and adopt Adam optimizer~\cite{kingma2014adam} with initial learning rate $10^{-4}$.
After 100k iterations, the learning rate is reduced to $10^{-5}$ until convergence.
To train Gaussian denoising models, we synthesize noisy videos by adding AWGN of $\sigma\in[0, 55]$ to clean ones.
For CRVD~\cite{yue2020supervised} dataset, as each training sequence only contains 7 frames, we mirror the training sequences to 14 frames to facilitate the model training, and crop patches with Bayer pattern preserving~\cite{liu2019learning}.
The evaluation is conducted on an RTX2080Ti.

\subsection{Ablation Study}
%
%
We conduct the ablation study on the Set8 dataset~\cite{tassano2019dvdnet} to assess the effectiveness of the proposed look-ahead recurrent module as well as its major components.

\begin{table}[t]
\scriptsize
\caption{Quantitative comparison of PSNR/SSIM on the Set8 dataset~\cite{tassano2019dvdnet} for Gaussian denoising. Hereinafter, {\color{red} red} and {\color{blue} blue} indicate the best and the second best results, respectively.}
\centering
\begin{tabular}{cccccccc}
\toprule
\textbf{Set8} & VBM4D\cite{maggioni2012video} & VNLB\cite{arias2018video} & DVDNet\cite{tassano2019dvdnet} & FastDVD\cite{tassano2020fastdvdnet} & VNLNet\cite{davy2021video} & PaCNet\cite{vaksman2021patch} & FloRNN\\
\midrule
$\sigma=10$ & 36.05/- & 37.26/- & 36.08/.9510 & 36.44/.9540 & \textcolor{blue}{37.28}/\textcolor{blue}{.9606} & 37.06/.9590 & \textcolor{red}{37.57}/\textcolor{red}{.9639}\\
$\sigma=20$ & 32.19/- & 33.72/- & 33.49/.9182 & 33.43/.9196 & \textcolor{blue}{34.02}/\textcolor{blue}{.9273} & 33.94/.9247 & \textcolor{red}{34.67}/\textcolor{red}{.9379}\\
$\sigma=30$ & 30.00/- & 31.74/- & 31.68/.8862 & 31.68/.8889 & -           & \textcolor{blue}{32.05}/\textcolor{blue}{.8921} & \textcolor{red}{32.97}/\textcolor{red}{.9138}\\
$\sigma=40$ & 28.48/- & 30.39/- & 30.46/.8564 & 30.46/.8608 & \textcolor{blue}{30.72}/.8622 & 30.70/\textcolor{blue}{.8623} & \textcolor{red}{31.75}/\textcolor{red}{.8911}\\
$\sigma=50$ & 27.33/- & 29.24/- & 29.53/.8289 & 29.53/\textcolor{blue}{.8351} & -           & \textcolor{blue}{29.66}/.8349 & \textcolor{red}{30.80}/\textcolor{red}{.8696}\\
avg         & 30.81/- & 32.47/- & 32.29/.8881 & 32.31/.8917 & -           & \textcolor{blue}{32.68}/\textcolor{blue}{.8946} & \textcolor{red}{33.55}/\textcolor{red}{.9153}\\
\bottomrule
\end{tabular}
\label{tab:set8}
\end{table}
\begin{table}[t]
\scriptsize
\caption{Quantitative comparison of PSNR/SSIM on the DAVIS dataset~\cite{pont20172017} for Gaussian denoising.}
\centering
\setlength{\tabcolsep}{0.4mm}
\begin{tabular}{cccccccc}
\toprule
\textbf{DAVIS} & VBM4D\cite{maggioni2012video} & VNLB\cite{arias2018video} & DVDNet\cite{tassano2019dvdnet} & FastDVD\cite{tassano2020fastdvdnet} & VNLNet\cite{davy2021video} & PaCNet\cite{vaksman2021patch} & FloRNN\\
\midrule
$\sigma=10$ & 37.58/- & 38.85/- & 38.13/.9657 & 38.71/.9672 & 39.56/.9707 & \textcolor{blue}{39.97}/\textcolor{blue}{.9713} & \textcolor{red}{40.16}/\textcolor{red}{.9755}\\
$\sigma=20$ & 33.88/- & 35.68/- & 35.70/.9422 & 35.77/.9405 & 36.53/.9464 & \textcolor{blue}{37.10}/\textcolor{blue}{.9470} & \textcolor{red}{37.52}/\textcolor{red}{.9564}\\
$\sigma=30$ & 31.65/- & 33.73/- & 34.08/.9188 & 34.04/.9167 & -           & \textcolor{blue}{35.07}/\textcolor{blue}{.9211} & \textcolor{red}{35.89}/\textcolor{red}{.9440}\\
$\sigma=40$ & 30.05/- & 32.32/- & 32.86/.8962 & 32.82/.8949 & 33.32/\textcolor{blue}{.8996} & \textcolor{blue}{33.57}/.8969 & \textcolor{red}{34.66}/\textcolor{red}{.9286}\\
$\sigma=50$ & 28.80/- & 31.13/- & 31.85/.8745 & 31.86/\textcolor{blue}{.8747} & -           & \textcolor{blue}{32.39}/.8743 & \textcolor{red}{33.67}/\textcolor{red}{.9131}\\
avg         & 32.39/- & 34.34/- & 34.52/.9195 & 34.64/.9188 & -           & \textcolor{blue}{35.62}/\textcolor{blue}{.9221} & \textcolor{red}{36.38}/\textcolor{red}{.9435}\\
\bottomrule
\end{tabular}
\label{tab:davis}
\end{table}
\begin{table}[t]
\scriptsize
\caption{Model complexity comparison on Set8 dataset~\cite{tassano2020fastdvdnet}.}
    \centering
    \setlength{\tabcolsep}{0.2mm}
    \begin{tabular}{lccccccc}
        \toprule
        & VBM4D\cite{maggioni2012video} & VNLB\cite{arias2018video} & DVDNet\cite{tassano2019dvdnet} & FastDVD\cite{tassano2020fastdvdnet} & VNLNet\cite{davy2021video} & PaCNet\cite{vaksman2021patch} & FloRNN\\
        \midrule
        PSNR & 30.81 & 32.47 & 32.29 & 32.31 & - & 32.68 & 33.55 \\
        \#.Frame & 13 & 13 & 5 & 5 & 15 & 15 & T/2 + 3 \\
        \#.Param(M) & - & - & 1.33 & 2.48 & 4.52 & 2.87 & 11.82 \\
        FLOPS(G) & - & - & 1231.8 & 661.8 & 2.02$\times 10^5$ & 6.13$\times 10^4$ & 3002.8 \\
        Time(s) & 420.0 & 156.0 & 2.51 & 0.08 & 1.65 & 35.24 & 0.31\\
        \bottomrule
    \end{tabular}
\label{tab:complexity}
\end{table}
\begin{figure}[h]
\centering
\begin{subfigure}[b]{0.24\linewidth}
    \centering
    \includegraphics[width=\linewidth]{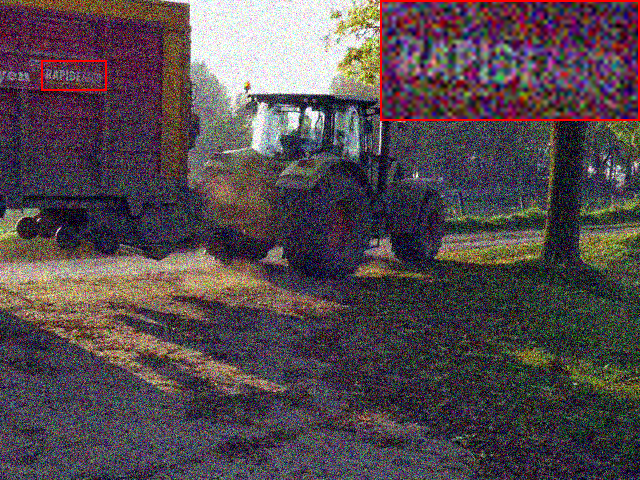}
    \caption*{\scriptsize Noisy($\sigma$=40)}
\end{subfigure}
\hfill
\begin{subfigure}[b]{0.24\linewidth}
    \centering
    \includegraphics[width=\linewidth]{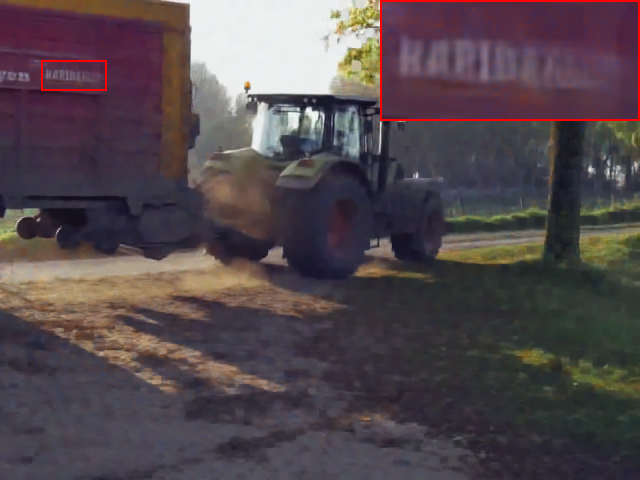}
    \caption*{\scriptsize VBM4D}
\end{subfigure}
\hfill
\begin{subfigure}[b]{0.24\linewidth}
    \centering
    \includegraphics[width=\linewidth]{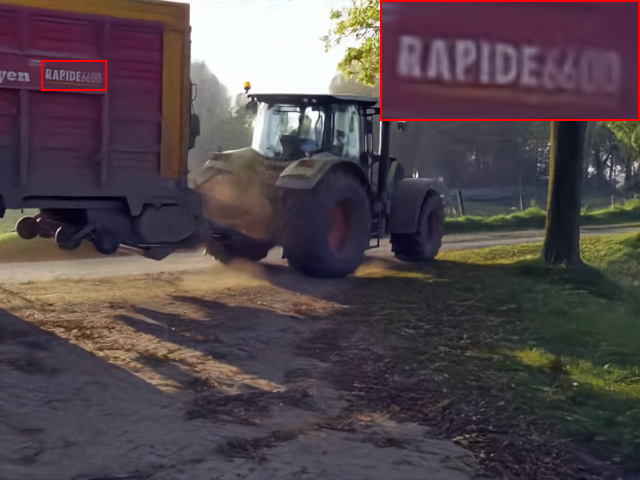}
    \caption*{\scriptsize DVDNet}
\end{subfigure}
\hfill
\begin{subfigure}[b]{0.24\linewidth}
    \centering
    \includegraphics[width=\linewidth]{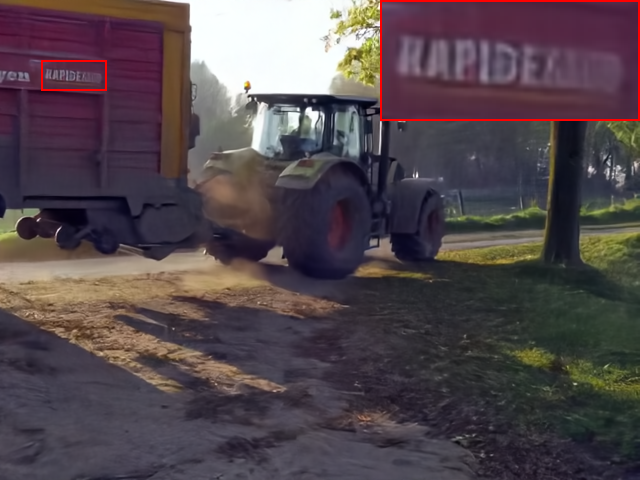}
    \caption*{\scriptsize FastDVDNet}
\end{subfigure}\\
\begin{subfigure}[b]{0.24\linewidth}
    \centering
    \includegraphics[width=\linewidth]{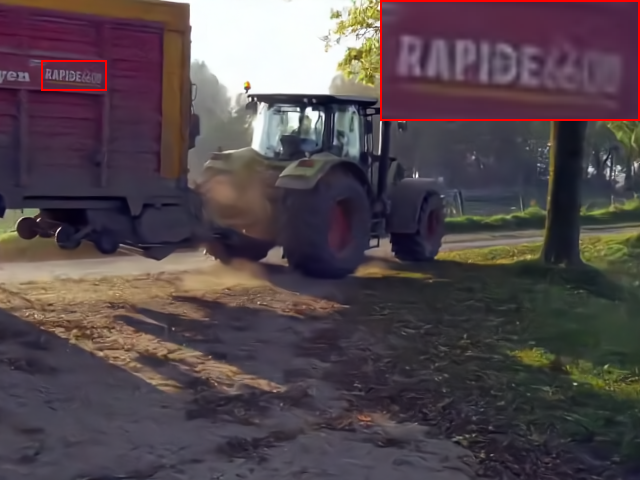}
    \caption*{\scriptsize VNLNet}
\end{subfigure}
\hfill
\begin{subfigure}[b]{0.24\linewidth}
    \centering
    \includegraphics[width=\linewidth]{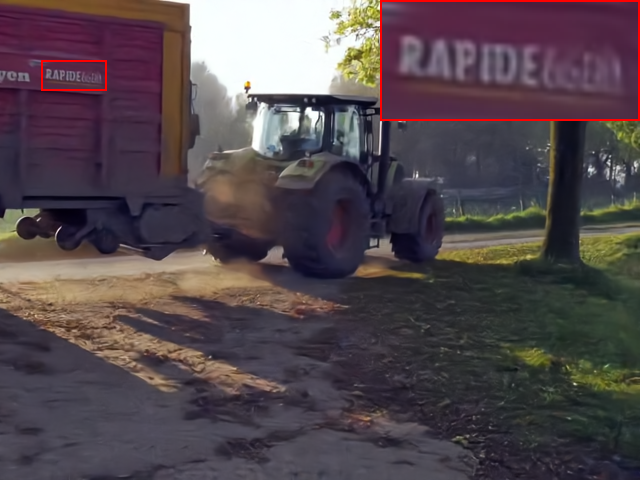}
    \caption*{\scriptsize PaCNet}
\end{subfigure}
\hfill
\begin{subfigure}[b]{0.24\linewidth}
    \centering
    \includegraphics[width=\linewidth]{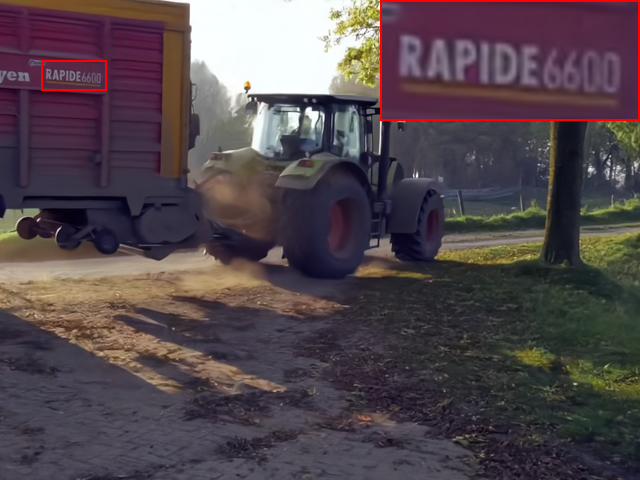}
    \caption*{\scriptsize FloRNN(Ours)}
\end{subfigure}
\hfill
\begin{subfigure}[b]{0.24\linewidth}
    \centering
    \includegraphics[width=\linewidth]{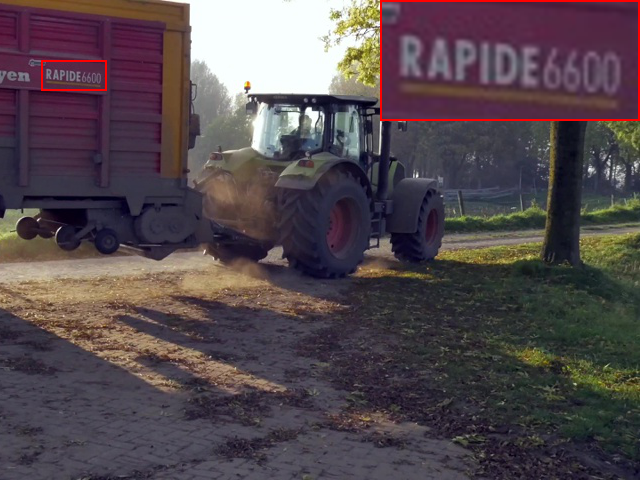}
    \caption*{\scriptsize GT}
\end{subfigure}

\caption{Visual comparison for Gaussian denoising ($\sigma$=40) on `tractor' sequence of the DAVIS dataset~\cite{pont20172017}.}
\label{fig:color_gaussian}
\end{figure}

\begin{table}[t]
\footnotesize
\caption{Quantitative comparison of PSNR/SSIM on CRVD dataset~\cite{yue2020supervised} for real-world raw video denoising.}
\centering
\scriptsize
\setlength{\tabcolsep}{1.4mm}
\begin{tabular}{cccccc}
\toprule
Method & RViDeNet\cite{yue2020supervised} & FastDVDNet\cite{tassano2020fastdvdnet} & EMVD\cite{maggioni2021efficient} & EDVR\cite{wang2019edvr} & FloRNN\\
\midrule
raw & 43.97/.9874 & 44.30/.9891 & 44.51/.9897 & \textcolor{blue}{44.71}/\textcolor{blue}{.9902} & \textcolor{red}{45.16}/\textcolor{red}{.9907}\\
sRGB & 39.95/.9792 & 39.91/.9812 & - & \textcolor{blue}{40.89}/\textcolor{blue}{.9838} & \textcolor{red}{41.01}/\textcolor{red}{.9843}\\
\bottomrule
\end{tabular}
\label{tab:crvd}
\end{table}
\begin{table}[t]
\scriptsize
\caption{Quantitative comparison of PSNR/SSIM on  IOCV dataset~\cite{kong2020comprehensive} for real-world video denoising in sRGB domain.}
\centering
\begin{tabular}{ccccccc}
\toprule
Method & Noisy & CVMSt-SVD\cite{kong2019color} & VBM4D\cite{maggioni2012video} & FastDVD\cite{tassano2020fastdvdnet} & VNLNet\cite{davy2021video} & FloRNN\\
\midrule
HUAWEI\_BC & 38.11/.9593 & 40.80/.9834 & 41.19/.9830 & 41.26/.9857 & \textcolor{blue}{41.35}/\textcolor{blue}{.9868} & \textcolor{red}{42.28}/\textcolor{red}{.9880}\\
HUAWEI\_FC & 38.58/.9413 & 38.71/.9780 & 38.76/.9785 & 38.03/.9776 & \textcolor{blue}{38.79}/\textcolor{blue}{.9809} & \textcolor{red}{39.57}/\textcolor{red}{.9828}\\
OPPO\_BC   & 32.06/.9071 & 33.44/.9508 & 33.26/.9456 & 33.05/.9476 & \textcolor{blue}{33.56}/\textcolor{blue}{.9544} & \textcolor{red}{33.75}/\textcolor{red}{.9545}\\
OPPO\_FC   & 36.90/.9447 & 39.66/.9791 & 39.56/.9785 & 39.06/.9751 & \textcolor{blue}{40.11}/\textcolor{red}{.9823} & \textcolor{red}{40.31}/\textcolor{blue}{.9821}\\
\bottomrule
\end{tabular}
\label{tab:iocv}
\end{table}

To demonstrate the effectiveness of FloRNN, we first compare it with two representative recurrent networks, \ie, ForwardRNN and BiRNN.
The main difference between the above three methods is how future frames are leveraged.
ForwardRNN takes the forward recurrent module to propagate information
from the first to the current frames, without leveraging any future information.
BiRNN deploys a backward recurrent module to propagate all future frames from the last frame to current frame.
Our FloRNN uses a look-ahead recurrent module for exploiting near-future frames.
Table~\ref{tab:ablation_architecture} and Fig.~\ref{fig:memory} show the quantitative comparison among them. 
From Table~\ref{tab:ablation_architecture}, FloRNN and BiRNN achieve better performance than ForwardRNN, witch demonstrates the importance of future information in video denoising.
Although BiRNN exhibits better quantitative results (0.19dB better than ours), it has a crucial issue that it can only be performed in an \emph{offline} manner.
As shown in Fig.~5, the memory consumption and latency of BiRNN grow linearly with respect to the sequence length. 
For example, denoising a 1000 frame sequence ($\sim$42s at 24$fps$), BiRNN consumes several hundred GB of memory, which limits its practicality for denoising long videos on common hardware.
In comparison, both ForwardRNN and our FloRNN are unidirectional algorithms with constant memory consumption and latency.
With the performance close to BiRNN, our FloRNN achieves much lower computation complexity and memory occupation, which demonstrates the effectiveness of our proposed look ahead recurrent module.

%

Moreover, we explore the effects of two components in the look-ahead module, \ie, (i) alignment mechanism, (ii) number of near-future frames $k$.
Quantitative results are listed in the Table~\ref{tab:ablation_components}.
Although backward warping is more popular in video restoration~\cite{tassano2019dvdnet, chan2021basicvsr, chan2021basicvsr++}, from Table~\ref{tab:ablation_components}(a), forward warping and backward warping achieve similar performance.
It can be explained that various splatting strategies~\cite{niklaus2020softmax} have been investigated to solve the conflict of target position in forward warping, and flow based feature warping in FloRNN may be robust to the hole problem of forward warping.
Besides, incorporating forward warping and border enlargement significantly reduces the amount of lost border pixels during alignment, and achieves 0.1dB performance gain.
Table~\ref{tab:ablation_components}(b) analyzes the effect of $k$.
When $k$ increases, the benefit from near-future information first increases then becomes steady. 
In our following experiments, we set $k$ = 3 for best performance and acceptable latency.
\subsection{Results}
We further compare our FloRNN with the state-of-the-art methods~\cite{maggioni2012video,arias2018video,tassano2019dvdnet,tassano2020fastdvdnet,davy2021video,vaksman2021patch,yue2020supervised,maggioni2021efficient,wang2019edvr,kong2019color} on both synthetic and real-world video denoising datasets. FloRNN shows appealing performance.

\noindent \textbf{Gaussian denoising on Set8 and DAVIS datasets.}
Table~\ref{tab:set8} and \ref{tab:davis} lists the quantitative comparison on Set8~\cite{tassano2019dvdnet} and DAVIS~\cite{pont20172017} datasets. 
From the table, our FloRNN outperforms the state-of-the-art methods by a large margin on both datasets in terms of PSNR/SSIM, which demonstrates its superiority on video denoising applications.
We also analyse the model complexity of each method in table~\ref{tab:complexity}, and found that FloRNN achieves good trade-off between performance and running time.
Specifically, FloRNN surpasses non-local based PaCNet~\cite{vaksman2021patch} by 0.87dB on Set8, while hundred times faster.
Compare to FastDVDNet~\cite{tassano2020fastdvdnet} which is especially suggested for efficiency, the performance gain of FloRNN is promoted to 1.24dB, with 4 times slower.
Experiments on DAVIS dataset shows similar results, which demonstrates that FloRNN generates best results with good efficiency.
Fig.~\ref{fig:color_gaussian} illustrates the qualitative comparison on a dynamic scene with noise level $\sigma$=40.
Despite the large motion and severe noise, FloRNN could clearly recover the characters, which are hard to recognize in other results.
More results are provided in the supplementary material.
\begin{figure}[t]
\centering
\begin{subfigure}[b]{0.159\linewidth}
    \centering
    \includegraphics[width=\linewidth]{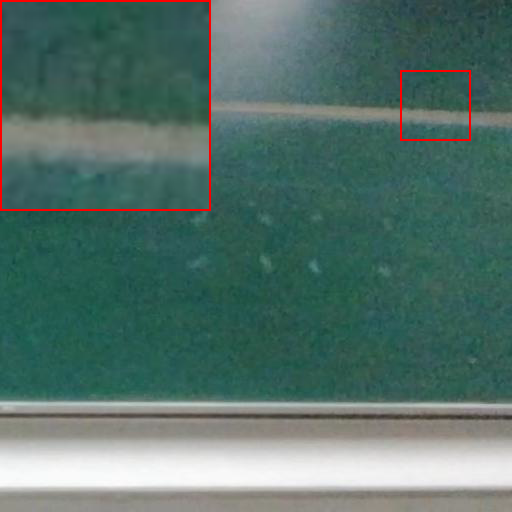}
    \caption*{\scriptsize Noisy}
\end{subfigure}
\hfill
\begin{subfigure}[b]{0.159\linewidth}
    \centering
    \includegraphics[width=\linewidth]{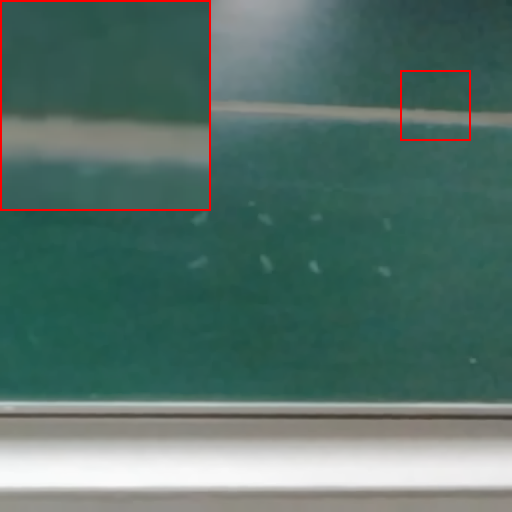}
    \caption*{\scriptsize VBM4D}
\end{subfigure}
\hfill
\begin{subfigure}[b]{0.159\linewidth}
    \centering
    \includegraphics[width=\linewidth]{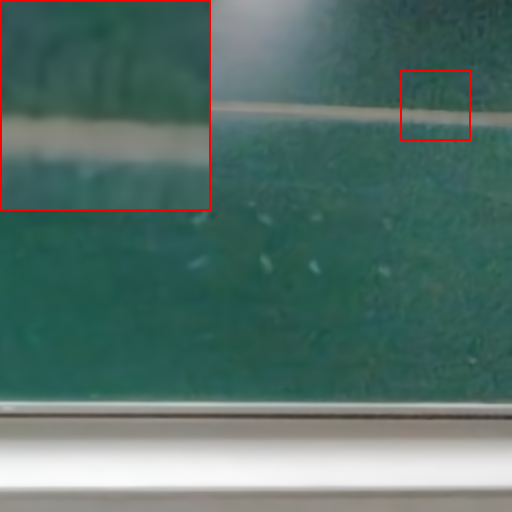}
    \caption*{\scriptsize FastDVDNet}
\end{subfigure}
\hfill
\begin{subfigure}[b]{0.159\linewidth}
    \centering
    \includegraphics[width=\linewidth]{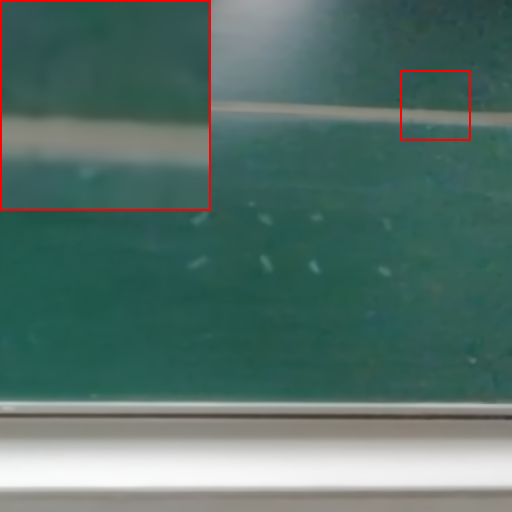}
    \caption*{\scriptsize VNLNet}
\end{subfigure}
\hfill
\begin{subfigure}[b]{0.159\linewidth}
    \centering
    \includegraphics[width=\linewidth]{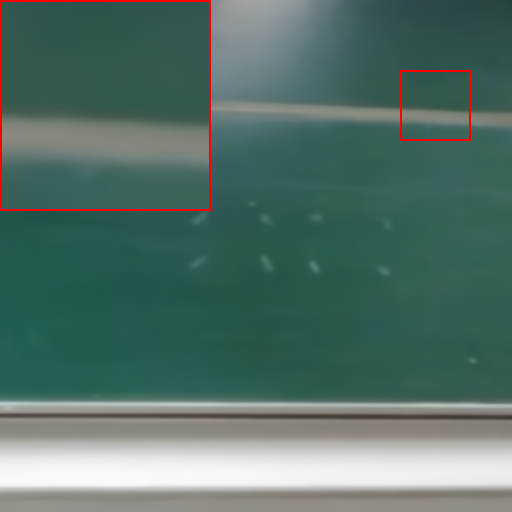}
    \caption*{\scriptsize FloRNN(Ours)}
\end{subfigure}
\hfill
\begin{subfigure}[b]{0.159\linewidth}
    \centering
    \includegraphics[width=\linewidth]{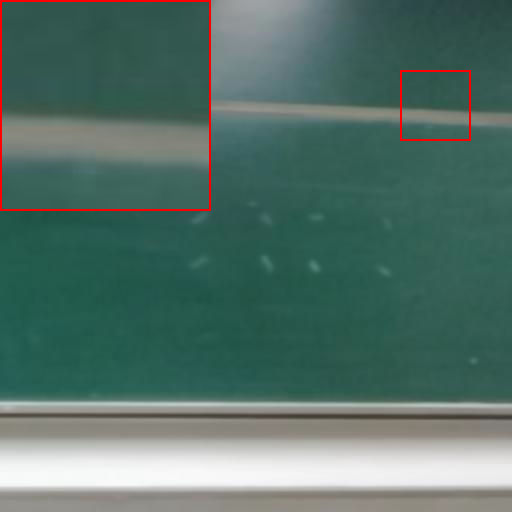}
    \caption*{\scriptsize GT}
\end{subfigure}
\caption{Visual comparison on the real-world IOCV dataset~\cite{kong2020comprehensive}.}
\label{fig:iocv}
\end{figure}
\begin{figure}[t]
\centering
\includegraphics[width=\textwidth]{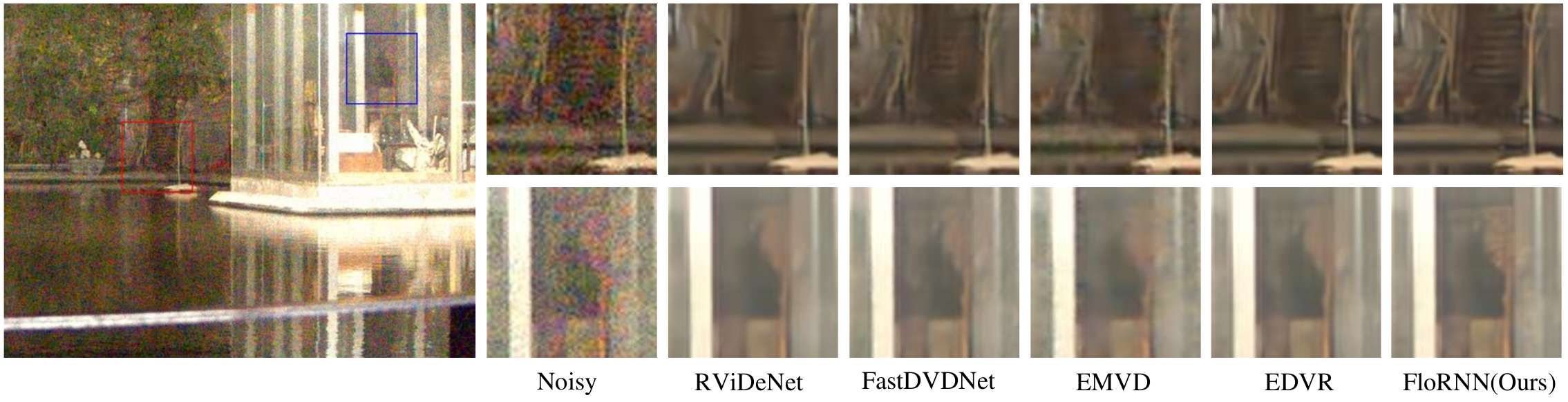}
\caption{Visual comparison of an outdoor scene on the CRVD dataset~\cite{yue2020supervised}, we render the results in raw domain to sRGB domain with a pretrained ISP.}
\label{fig:crvd}
\end{figure}

\noindent \textbf{Raw domain video denoising on CRVD dataset.}
We further evaluate our FloRNN on real-world video denoising datasets.
On CRVD dataset~\cite{yue2020supervised}, EMVD~\cite{maggioni2021efficient} adopts forward recurrent mechanism designed for mobile devices, but performs inferior due to lack of future information when scaling up on GPUs.
EDVR~\cite{wang2019edvr} utilizes pyramid deformable convolution for alignment and achieves good results, but its multi-frame convolutional backbone limits the long term propagation.
Instead, from Table~\ref{tab:crvd}, our FloRNN benefits both long term propagation of historical frames and near future information, and outperforms EDVR by 0.45dB on raw domain.
The performance on sRGB domain is calculated by rendering the raw results into sRGB ones with a pretrained ISP network.
Our performance gain decreases on sRGB domain, which is possibly due to the inaccuracy of the ISP module.

\noindent \textbf{sRGB domain video denoising on IOCV dataset.}
As IOCV~\cite{kong2020comprehensive} does not provide training sets, we apply our Gaussian denoising models for IOCV videos, and tune the noise level for each subset to get best results the same as~\cite{kong2020comprehensive}.
Nonetheless, FloRNN achieves favorable results and surpasses competing methods (see Table~\ref{tab:iocv}).
Visual comparison in Fig.~\ref{fig:iocv} and Fig.~\ref{fig:crvd} show that our FloRNN is able to remove the real-world video noise and recover fine-grained details.

\section{Conclusion}
In this paper, we propose a novel recurrent network for unidirectional video denoising.
Look-ahead recurrent module is sufficient to exploit near-future frames in a forward manner.
Combining with forward module, FloRNN achieves near BiRNN performance with constant memory occupation and latency.
By analyzing backward and forward warping mechanisms, we found incorporating forward warping and border enlargement is favorable to address the border information missing problem during alignment.
Experimental results demonstrate the superiority of the proposed method in removing both synthetic and real world noise.
Our unidirectional video denoising algorithm is beneficial to various video applications, \eg, video conference, live streaming, and could dynamically adjust the look-ahead step $k$ for balance denoising results and latency.

~\\
\noindent\textbf{Acknowledgement.} This work is partially supported by the National Natural Science Foundation of China (NSFC) under Grant No.s 62006064 and U19A2073.
\clearpage
%
%
\bibliographystyle{splncs04}
\bibliography{egbib}

\renewcommand\thesection{\Alph{section}}
\renewcommand\thesubsection{\thesection.\arabic{subsection}}
\renewcommand\thefigure{\Alph{figure}}
\renewcommand\thetable{\Alph{table}}
\setcounter{section}{0}
\setcounter{subsection}{0}
\setcounter{figure}{0}
\setcounter{table}{0}
\section{Quantitative results of BasicVSR~\cite{chan2021basicvsr} and BasicVSR++~\cite{chan2021basicvsr++}}
In addition to our implemented BiRNN, BasicVSR~\cite{chan2021basicvsr} and BasicVSR++~\cite{chan2021basicvsr++} are two popular BiRNN methods for video restoration.
BasicVSR~\cite{chan2021basicvsr} is introduced as a strong baseline with essential components for video super-resolution.
BasicVSR++~\cite{chan2021basicvsr++} further improves BasicVSR~\cite{chan2021basicvsr} in propagation and alignment, and generalizes to compressed video enhancement~\cite{Yang_2021_CVPR}.
However, their performance on video denoising is not well investigated.
In this section, we conduct experiments to validate these two methods on video denoising task.
\begin{table}[h]
\scriptsize
\caption{Quantitative comparison (PSNR/SSIM) of BasicVSR~\cite{chan2021basicvsr} and BasicVSR++~\cite{chan2021basicvsr++} for video denoising task on Set8 dataset~\cite{tassano2019dvdnet}.}
    \centering
    \begin{tabular}{c|c|c|c|c}
        \hline
        \textbf{Set8} & \multicolumn{2}{c|}{BasicVSR~\cite{chan2021basicvsr}} & \multicolumn{2}{c}{BasicVSR++~\cite{chan2021basicvsr++}} \\
        \hline
        downsample/upsample & \CheckmarkBold & \XSolidBrush & \CheckmarkBold & \XSolidBrush\\
        \hline
        $\sigma=10$ & 35.88/.9453 & 37.78/.9635 & 36.66/.9548 & 37.92/.9643\\
        $\sigma=20$ & 32.80/.9046 & 34.92/.9386 & 33.85/.9256 & 35.18/9408\\
        $\sigma=30$ & 31.02/.8694 & 33.24/.9159 & 32.24/.8996 & 33.56/.9195\\
        $\sigma=40$ & 29.78/.8379 & 32.03/.8945 & 31.10/.8759 & 32.40/.8996\\
        $\sigma=50$ & 28.83/.8091 & 31.07/.8738 & 30.20/.8539 & 31.48/.8808\\
        avg & 31.66/.8733 & 33.81/.9173 & 32.81/.9020 & 34.11/.9210\\
        \hline
        Time(s) & 0.06 & 0.65 & 0.08 & 1.84\\
        \hline
    \end{tabular}
    \label{tab:basicvsr++}
\end{table}

BasicVSR~\cite{chan2021basicvsr} and BasicVSR++~\cite{chan2021basicvsr++} are originally suggested for video super-resolution task, so they use pixel-shuffle~\cite{shi2016real} upsample layers at the end of the network to increase the spatial resolution.
But for video denoising task, the upsample layers are no longer needed.
There are two solutions to fit VSR networks to video denoising task: introducing additional downsample layers at the beginning of the network~\cite{chan2022generalization}, or just removing the upsample layers.
The former solution allows the network computing on low resolution video features, which seems as a more efficient choice.
But we found the downsample and upsample layers are very harmful to video denoising performance.

We use the official code~\footnote{https://github.com/open-mmlab/mmediting} to train the BasicVSR and BasicVSR++ on video denoising task with two variants: one add downsample layers at the beginning of the network~\cite{chan2022generalization}, the other remove the upsample layers at the end of the network.
We set the training video length to 10 instead of 30~\cite{chan2022generalization} due to GPU memory limit, so the results of downsample/upsample version BasicVSR++ is slightly lower ($\sim$0.2dB) than reported in~\cite{chan2022generalization}.
From table~\ref{tab:basicvsr++}, we found that downsample/upsample layers affect the denoising performance.
Although 4$\times$ downsample the input video and restoring in the low resolution feature space can reduce the computing complexity and speed up the inference, the performance drop can be up to 1$\sim$2 dB.
This indicates that getting rid of downsample/upsample layers is essential for best performing video denoising networks.

\section{Extend FloRNN to BasicVSR++~\cite{chan2021basicvsr++}}
\begin{table}[t]
\scriptsize
    \centering
    \caption{Quantitative comparison of PSNR with BasicVSR++~\cite{chan2021basicvsr++} on Set8 dataset~\cite{tassano2019dvdnet}. We apply flow-guided deformable alignment and second order propagation to FloRNN, named FloRNN++.}
    \begin{tabular}{c|c|c|c|c|c|c|c}
        \hline
         & BiRNN & (A) & (B) & (C) & BasicVSR++~\cite{chan2021basicvsr++} & FloRNN & FloRNN++\\
        \hline
        Flow-Guided Deform. Align & & \CheckmarkBold & & & \CheckmarkBold & & \CheckmarkBold\\
        Second-Order Propagation & & & \CheckmarkBold & & \CheckmarkBold & & \CheckmarkBold\\
        Grid Propagation & & & & \CheckmarkBold & \CheckmarkBold & &\\
        \hline
        PSNR & 33.74 & 33.86 & 33.79 & 33.84 & 34.11 & 33.55 & 33.72\\
        \hline
    \end{tabular}
    \label{tab:ablation_basicvsr_plusplus}
\end{table}
In this subsection, we show FloRNN can benefit from improvements of state-of-the-art recurrent methods.
We extend FloRNN to a state-of-the-art BiRNN method, \ie, BasicVSR++~\cite{chan2021basicvsr++}.
BasicVSR++~\cite{chan2021basicvsr++} improves BasicVSR with three modules, \ie, second-order propagation, grid propagation and flow-guided deformable alignment.
Due to grid propagation performs bidirectional propagation twice, it can not be equipped to FloRNN.
So we only apply second-order propagation and flow-guided deformable alignment to our FloRNN, named as FloRNN++.
From table~\ref{tab:ablation_basicvsr_plusplus}, with the aid of two improvements, FloRNN++ outperforms 0.17dB over FloRNN, which is comparable with our implemented BiRNN.
This demonstrates FloRNN can keep up with advances in BiRNN methods.
Although BasicVSR++~\cite{chan2021basicvsr++} achieves better quantitative results, it suffers the common issue of BiRNNs, \ie, large memory consumption, long latency and can only be performed in an offline manner.
In contrast, with the proposed look-ahead module, our FloRNN can address the offline issue and be applied to various real-time applications.

\section{Knowledge Distillation}
Analogous to other video denoising networks, our FloRNN can be simply trained from scratch using the reconstruction loss,
\begin{equation}
\mathcal{L}_{rec} = \sum_{t=1}^{T}(\hat{\mathbf{x}}_t - \mathbf{x}_t)^2, 
\end{equation}
where $T$ denotes the number of video frames.
\begin{table}[t]
\footnotesize
\setlength{\tabcolsep}{1em}
\centering
\caption{Ablation study of knowledge distillation on Set8 dataset~\cite{tassano2019dvdnet}, models with and without knowledge distillation show comparable results, which indicates our $F_l$ is able to mimic the $F_b$ of BiRNN and learn feature complementary to $F_f$.}
\label{tab:distill}
\begin{tabular}{ccc}
    \toprule
    Knowledge Distillation & \XSolidBrush & \CheckmarkBold\\
    \midrule
    PSNR/SSIM & 33.55/0.9153 & 33.53/0.9061\\
    \bottomrule
\end{tabular}
\end{table}
\begin{figure}[t]
\centering
\begin{subfigure}[b]{0.32\linewidth}
\centering
\includegraphics[width=\linewidth]{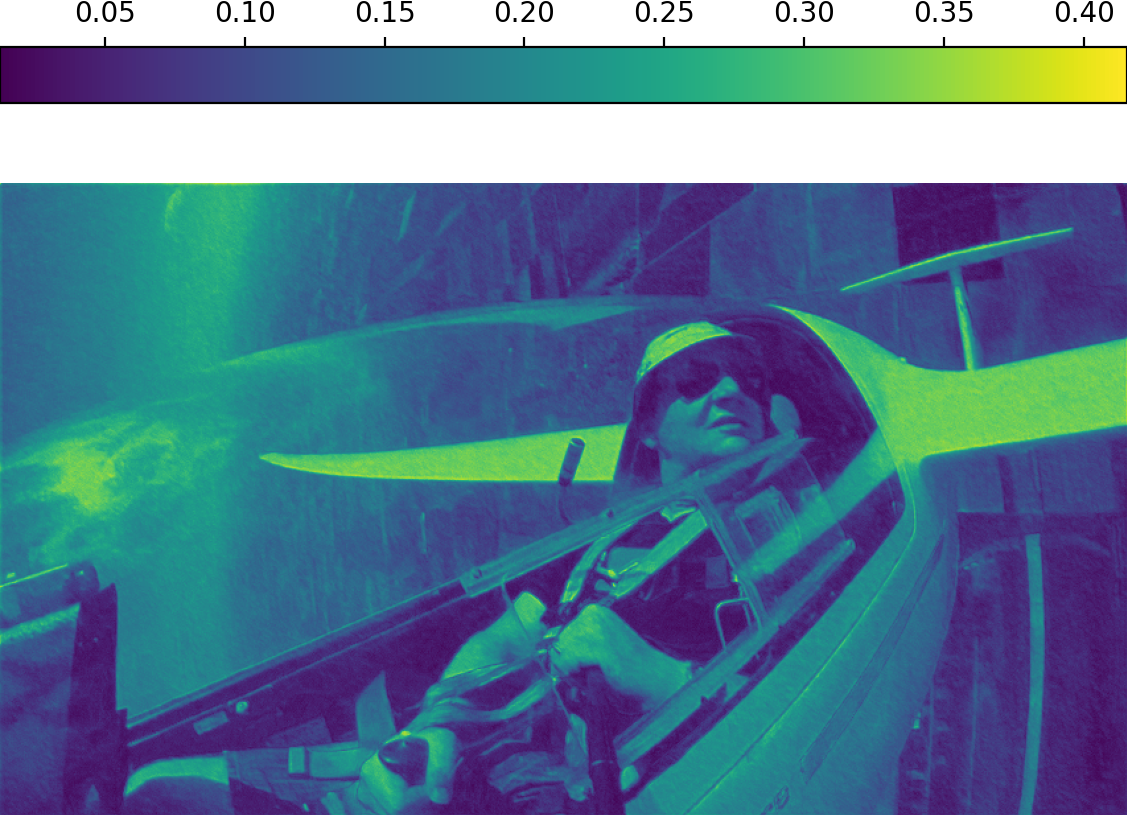}
\caption*{$h^f_t$}
\end{subfigure}
\hfill
\begin{subfigure}[b]{0.32\linewidth}
\centering
\includegraphics[width=\linewidth]{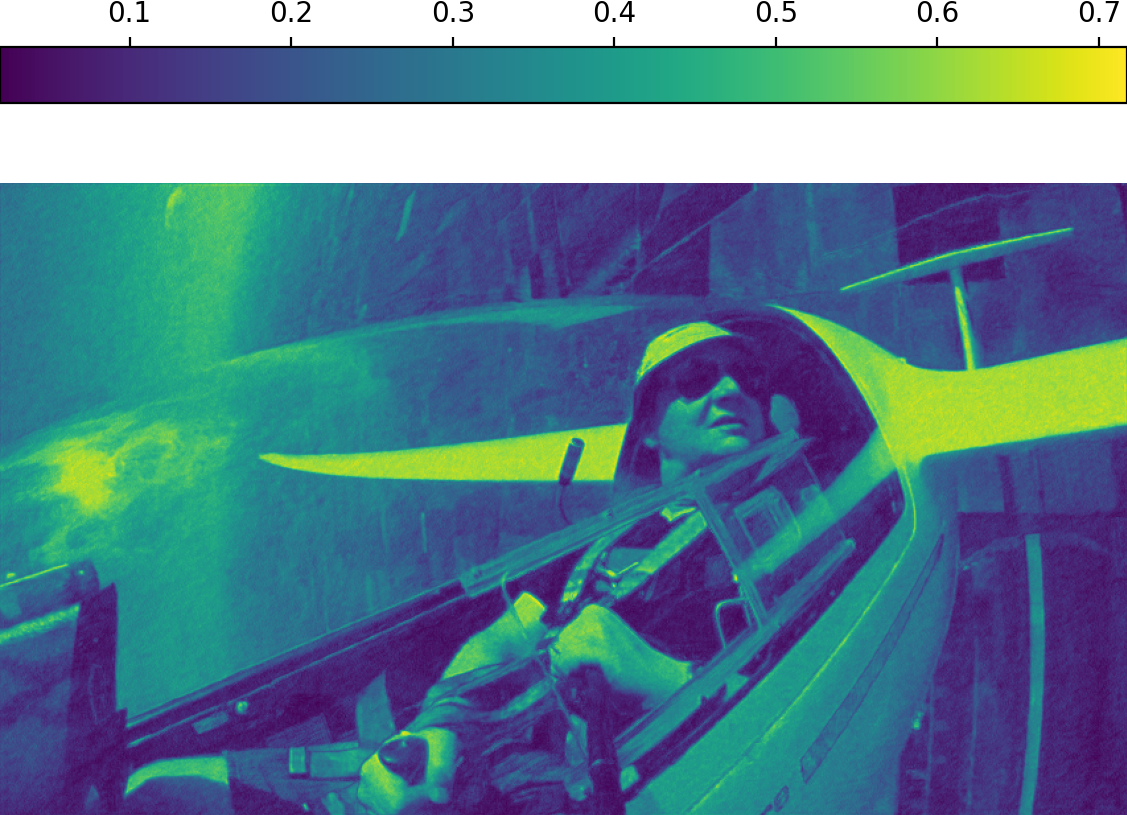}
\caption*{$h^f_t$}
\end{subfigure}
\hfill
\begin{subfigure}[b]{0.32\linewidth}
\centering
\includegraphics[width=\linewidth]{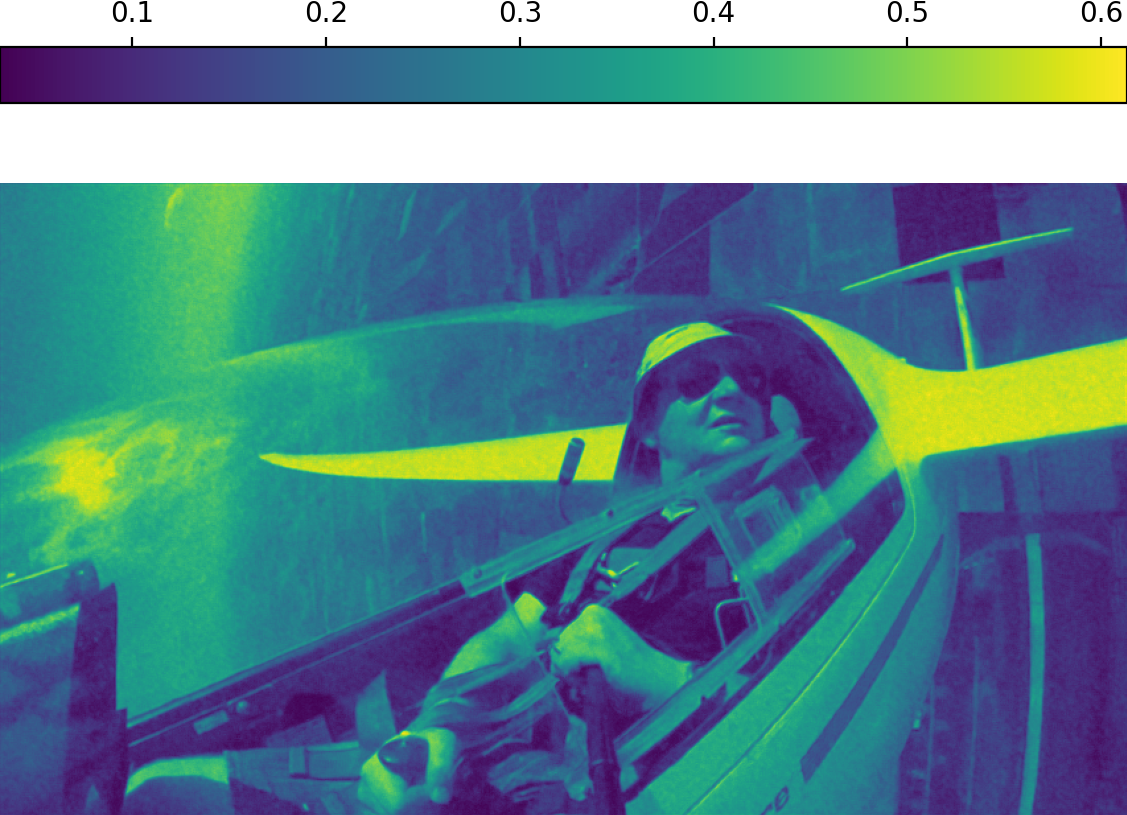}
\caption*{$h^f_t$}
\end{subfigure}\\
\begin{subfigure}[b]{0.32\linewidth}
\centering
\includegraphics[width=\linewidth]{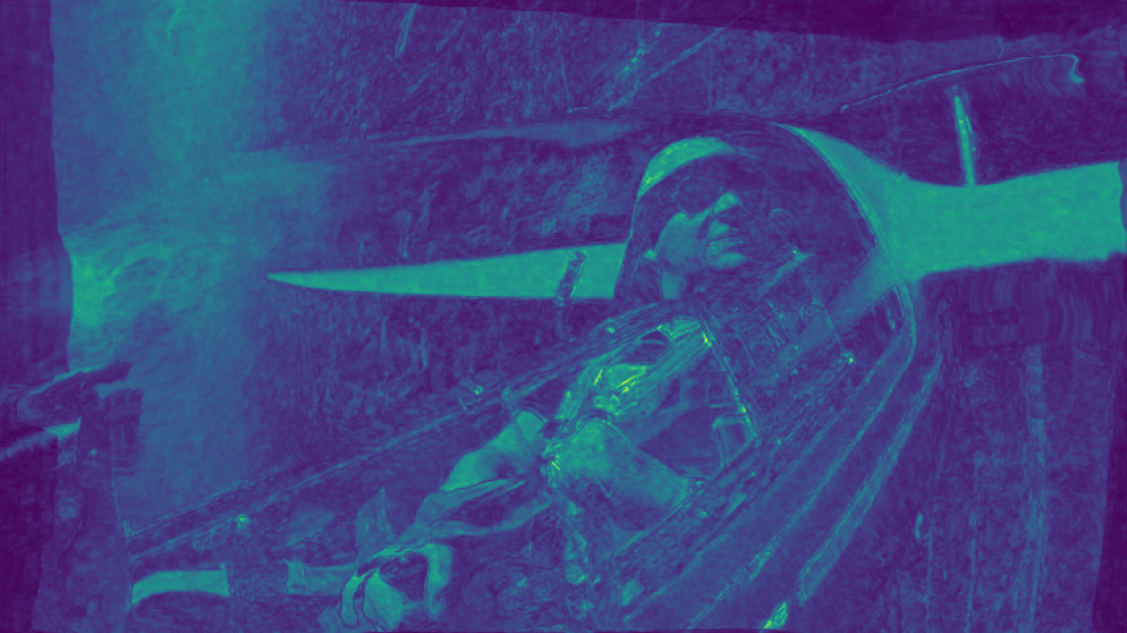}
\caption*{$h^l_{t+k\rightarrow t}$}
\end{subfigure}
\hfill
\begin{subfigure}[b]{0.32\linewidth}
\centering
\includegraphics[width=\linewidth]{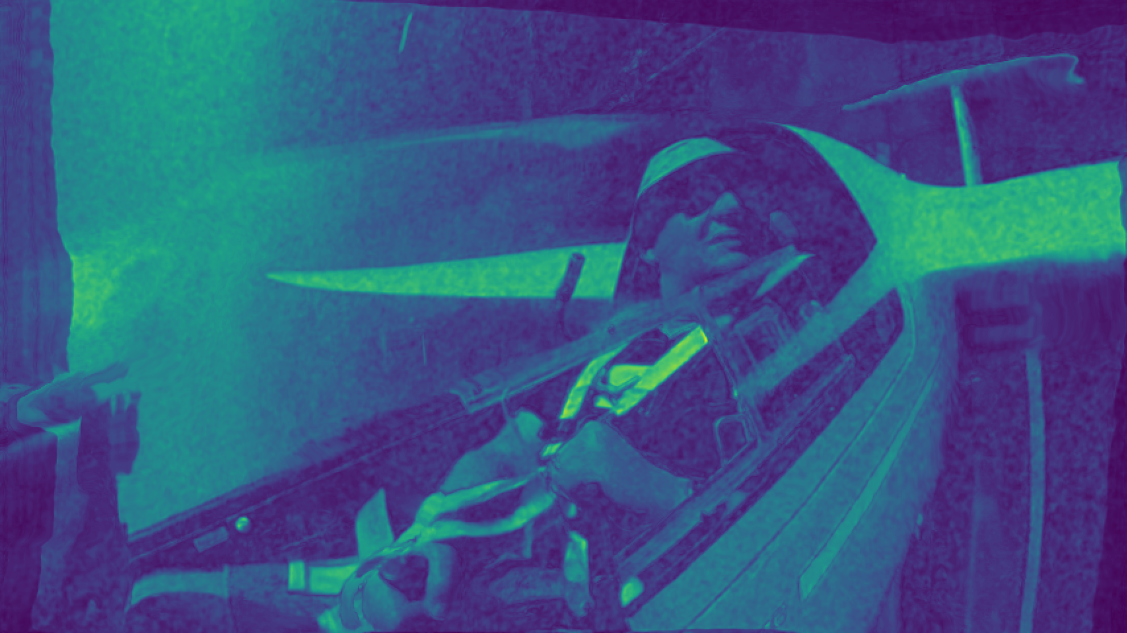}
\caption*{$h^l_{t+k\rightarrow t}$}
\end{subfigure}
\hfill
\begin{subfigure}[b]{0.32\linewidth}
\centering
\includegraphics[width=\linewidth]{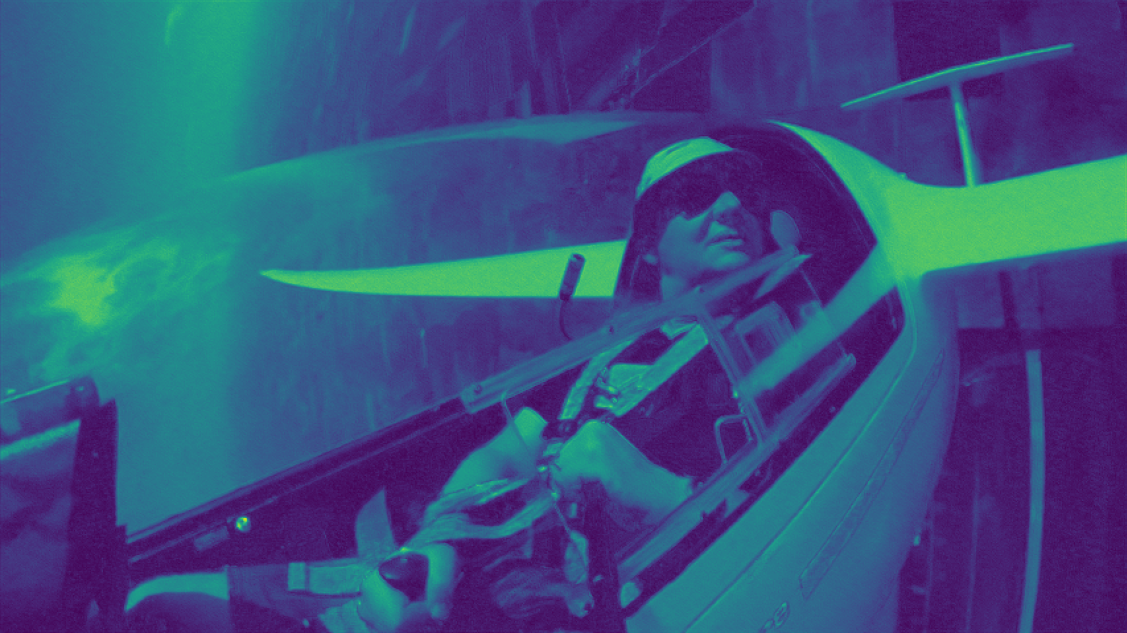}
\caption*{$h^b_t$}
\end{subfigure}\\
\begin{subfigure}[b]{0.32\linewidth}
\centering
\caption{FloRNN}
\end{subfigure}
\hfill
\begin{subfigure}[b]{0.32\linewidth}
\centering
\caption{FloRNN+distillation}
\end{subfigure}
\hfill
\begin{subfigure}[b]{0.32\linewidth}
\centering
\caption{BiRNN}
\end{subfigure}
\caption{Visual comparision of hidden features. Look-ahead feature with knowledge distillation (b) is more similar to backward feature of BiRNN (c), in comparison to the no distillation counterpart (a). The features are visualized with their $L_{\infty}$ norm.}
\label{fig:distill}
\end{figure}

Nonetheless, our FloRNN shares many similarities with BiRNN. 
They both adopt a forward recurrent module and a decoder. 
The look-ahead recurrent module in FloRNN is suggested to play a similar role as the backward recurrent module in BiRNN for leveraging information from future frames.
In order to show the feasibility of look-ahead recurrent module in mimicking backward recurrent module, we further suggest an alternative training scheme by incorporating pre-trained BiRNN and distillation loss. 
Specifically, we first train a BiRNN with reconstruction loss.
Then we substitute the backward recurrent module of BiRNN with our look-ahead recurrent module.
And distillation loss is deployed to mimic the backward feature $\mathbf{h}^b_t$ with aligned look-ahead feature $\mathbf{h}^l_{t+k\rightarrow t}$,
\begin{equation}
\mathcal{L}_{distill} = \sum_{t=1}^{T}\left|\mathbf{h}^l_{t+k\rightarrow t} - \mathbf{h}^b_t\right|.
\end{equation}
Knowledge distillation encourages the look-ahead recurrent module to learn feature similar to the backward recurrent module in BiRNN.
And reconstruction loss is also used to finetune the look-ahead recurrent module and decoder.
From Fig.~\ref{fig:distill}, the look-ahead feature $h^l_{t+k\rightarrow t}$ of the knowledge distillation counterpart is similar to the backward feature $h^b_t$ of BiRNN.
As shown in Table~\ref{tab:distill}, we empirically find such scheme achieves comparable performance in comparison to training from scratch using $\mathcal{L}_{rec}$.
This indicates that look-ahead recurrent module is able to mimic backward recurrent module and learn hidden feature complementary to $F_f$ for video denoising.

\section{More Experimental Results}
\begin{figure*}[htbp]
\centering
\begin{subfigure}[b]{0.24\linewidth}
    \centering
    \includegraphics[width=\linewidth]{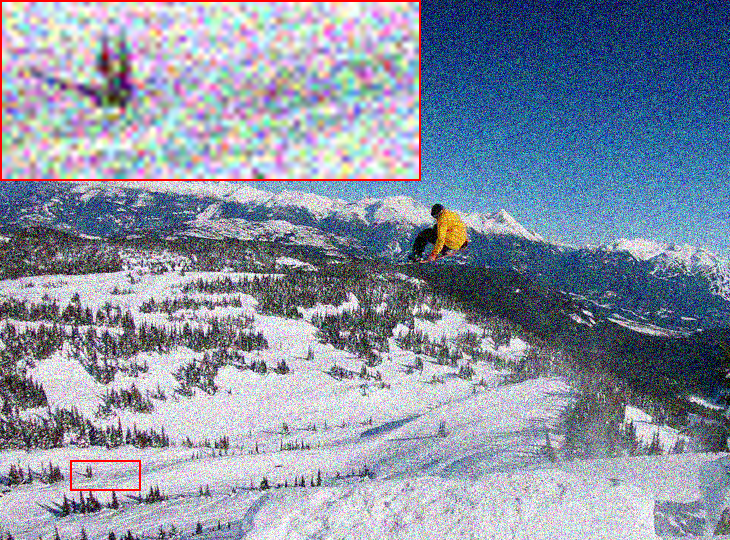}
    \caption{Noisy($\sigma$=40)}
\end{subfigure}
\hfill
\begin{subfigure}[b]{0.24\linewidth}
    \centering
    \includegraphics[width=\linewidth]{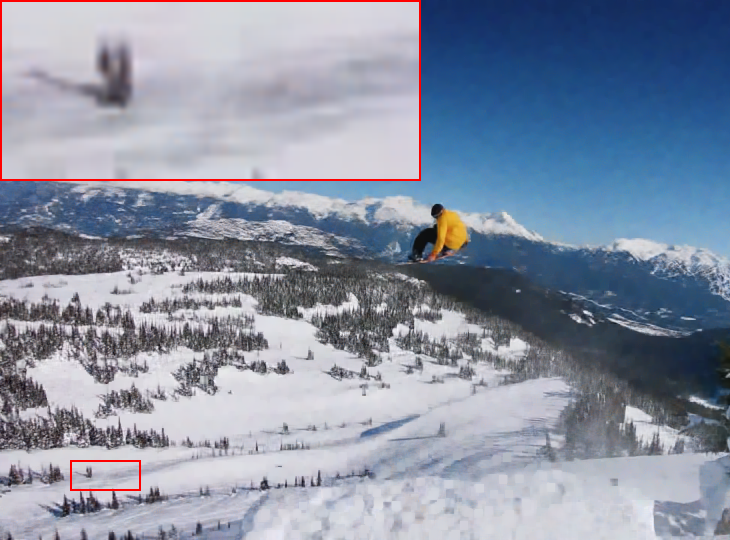}
    \caption{VBM4D}
\end{subfigure}
\hfill
\begin{subfigure}[b]{0.24\linewidth}
    \centering
    \includegraphics[width=\linewidth]{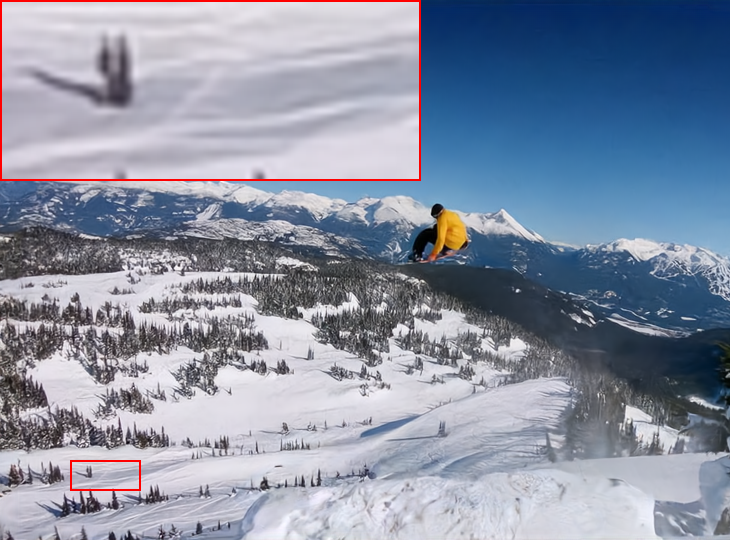}
    \caption{DVDNet}
\end{subfigure}
\hfill
\begin{subfigure}[b]{0.24\linewidth}
    \centering
    \includegraphics[width=\linewidth]{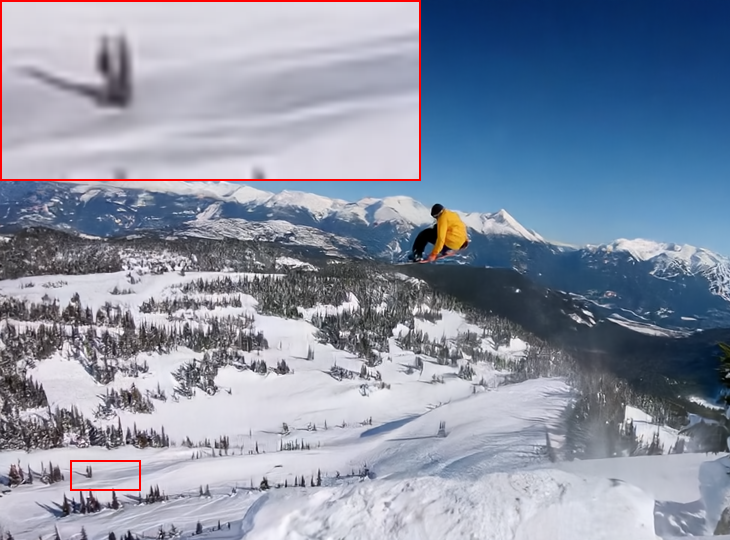}
    \caption{FastDVDNet}
\end{subfigure}\\
\begin{subfigure}[b]{0.24\linewidth}
    \centering
    \includegraphics[width=\linewidth]{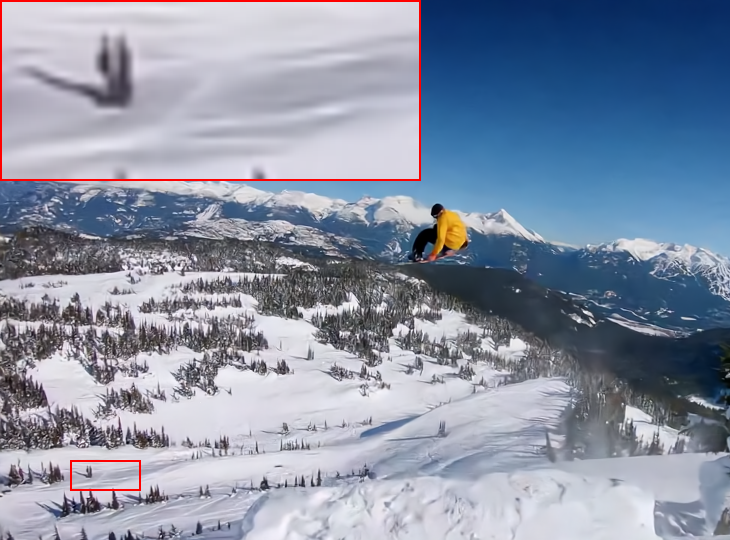}
    \caption{VNLNet}
\end{subfigure}
\hfill
\begin{subfigure}[b]{0.24\linewidth}
    \centering
    \includegraphics[width=\linewidth]{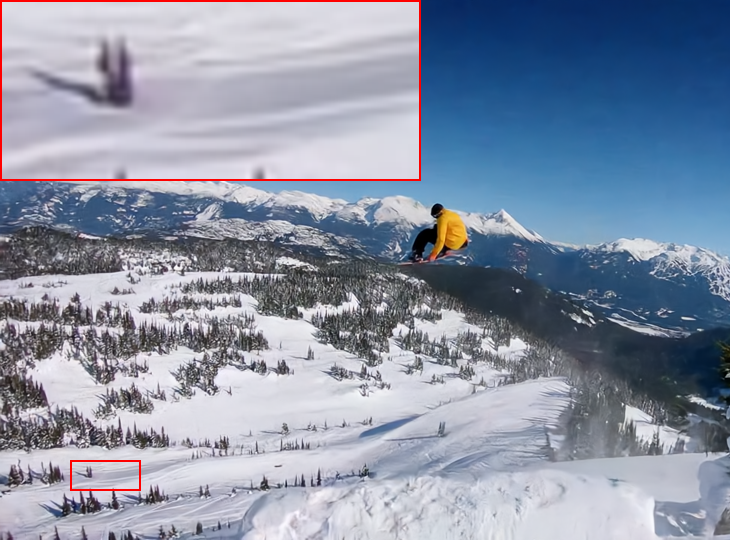}
    \caption{PaCNet}
\end{subfigure}
\hfill
\begin{subfigure}[b]{0.24\linewidth}
    \centering
    \includegraphics[width=\linewidth]{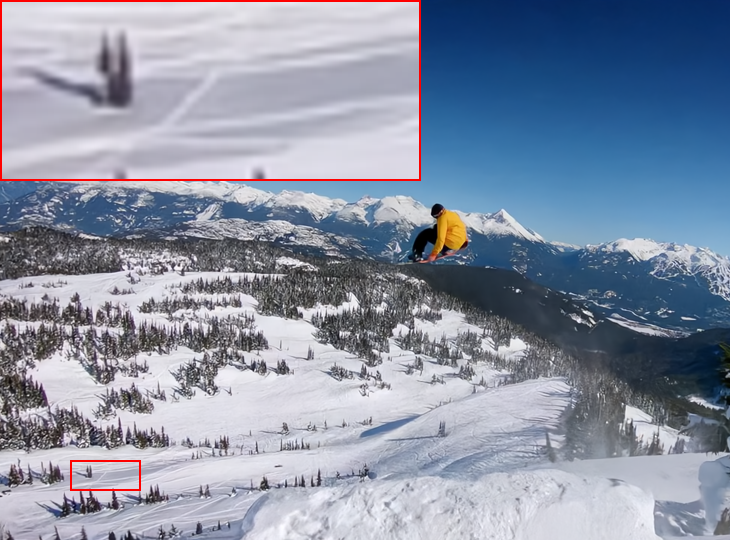}
    \caption{FloRNN(Ours)}
\end{subfigure}
\hfill
\begin{subfigure}[b]{0.24\linewidth}
    \centering
    \includegraphics[width=\linewidth]{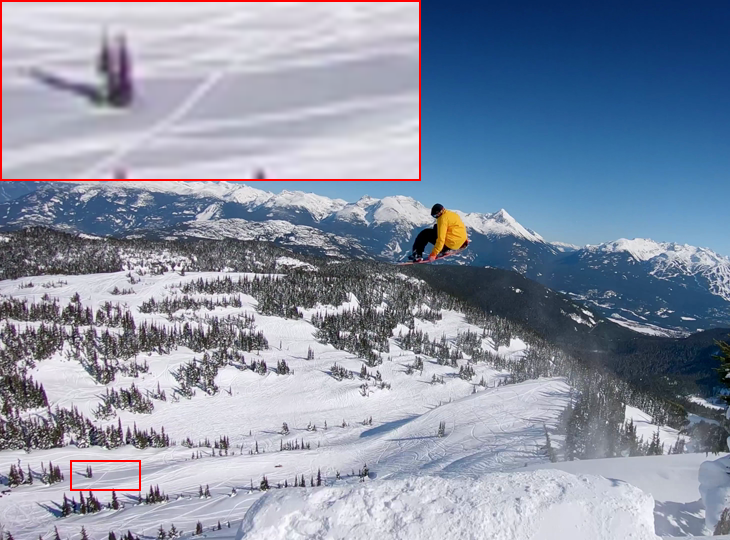}
    \caption{GT}
\end{subfigure}
\caption{More visual comparison for Gaussian denoising ($\sigma$ = 40) on the Set8 dataset~\cite{tassano2019dvdnet}.}
\label{fig:Set8}
\end{figure*}

\begin{figure*}[htbp]
\centering
\begin{subfigure}[b]{0.24\linewidth}
    \centering
    \includegraphics[width=\linewidth]{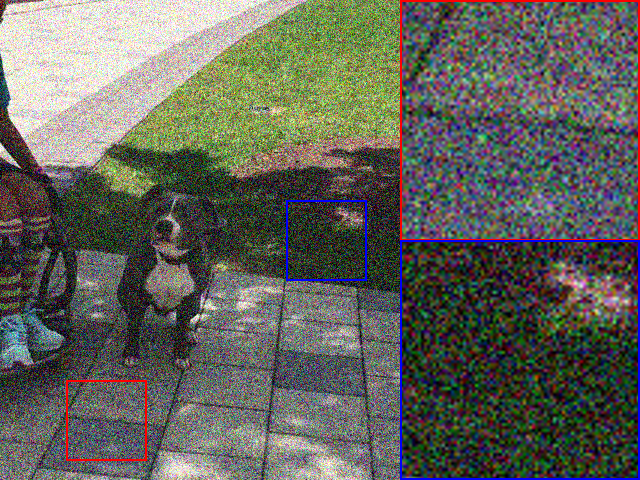}
    \caption{Noisy($\sigma$=40)}
\end{subfigure}
\hfill
\begin{subfigure}[b]{0.24\linewidth}
    \centering
    \includegraphics[width=\linewidth]{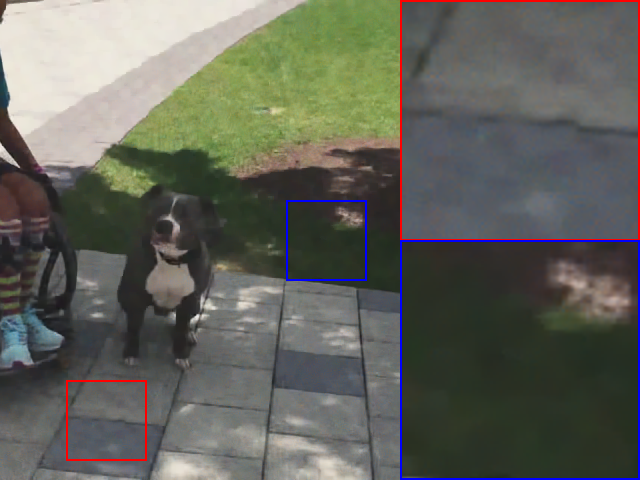}
    \caption{VBM4D}
\end{subfigure}
\hfill
\begin{subfigure}[b]{0.24\linewidth}
    \centering
    \includegraphics[width=\linewidth]{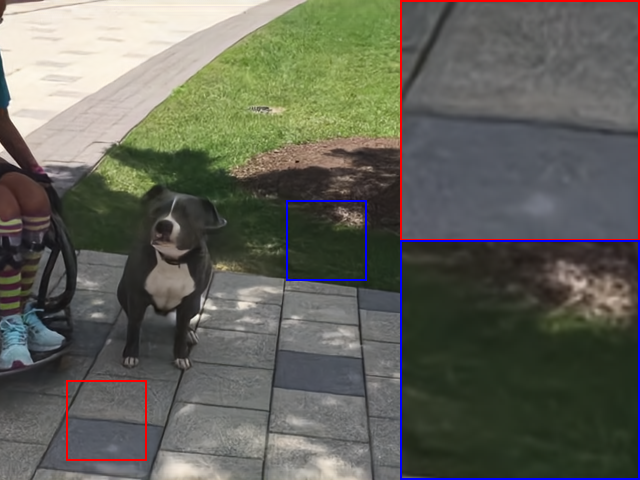}
    \caption{DVDNet}
\end{subfigure}
\hfill
\begin{subfigure}[b]{0.24\linewidth}
    \centering
    \includegraphics[width=\linewidth]{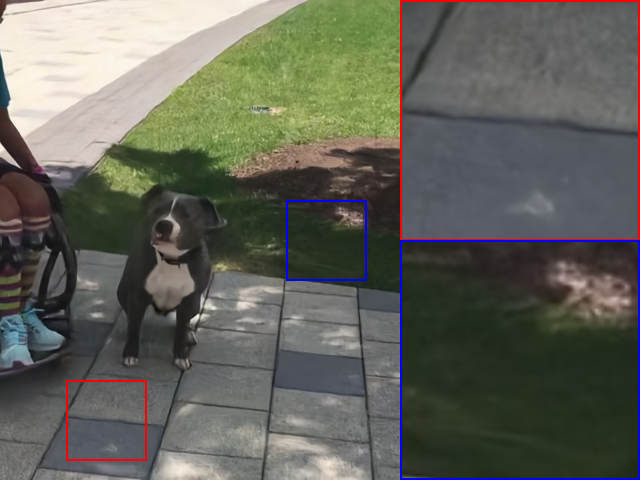}
    \caption{FastDVDNet}
\end{subfigure}\\
\begin{subfigure}[b]{0.24\linewidth}
    \centering
    \includegraphics[width=\linewidth]{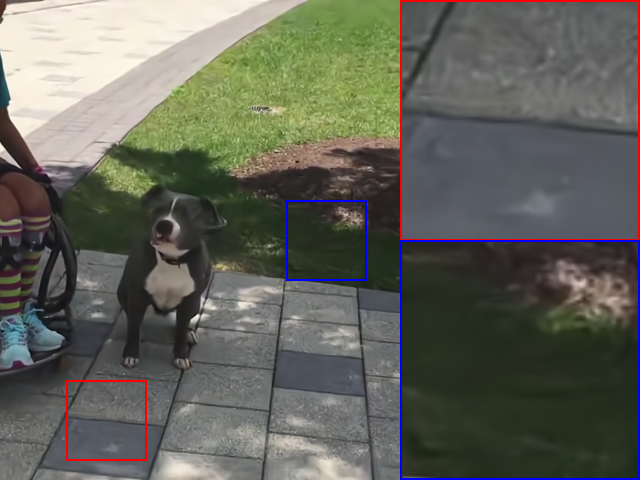}
    \caption{VNLNet}
\end{subfigure}
\hfill
\begin{subfigure}[b]{0.24\linewidth}
    \centering
    \includegraphics[width=\linewidth]{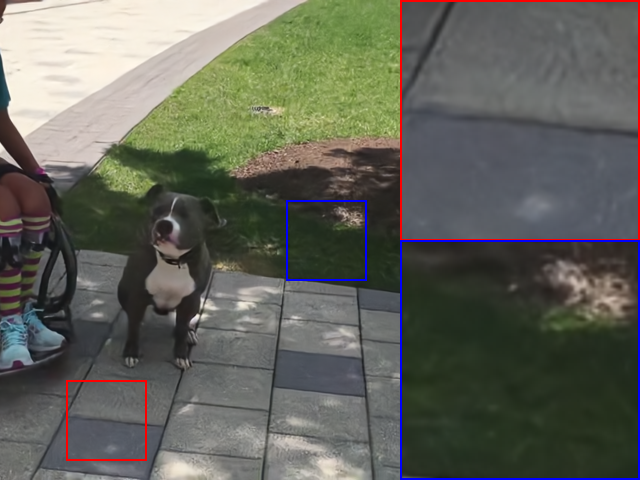}
    \caption{PaCNet}
\end{subfigure}
\hfill
\begin{subfigure}[b]{0.24\linewidth}
    \centering
    \includegraphics[width=\linewidth]{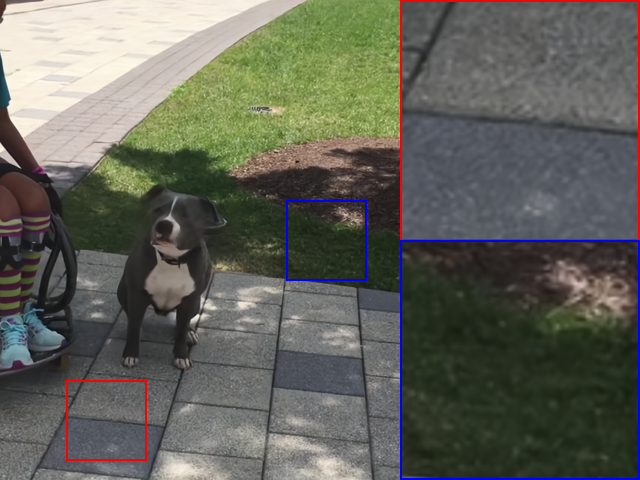}
    \caption{FloRNN(Ours)}
\end{subfigure}
\hfill
\begin{subfigure}[b]{0.24\linewidth}
    \centering
    \includegraphics[width=\linewidth]{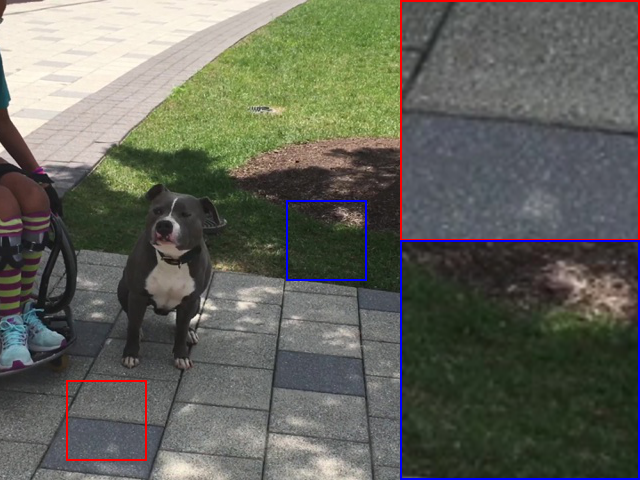}
    \caption{GT}
\end{subfigure}
\caption{More visual comparison for Gaussian denoising ($\sigma$ = 40) on the DAVIS dataset~\cite{pont20172017}.}
\label{fig:davis}
\end{figure*}

\begin{table}[t]
\footnotesize
\centering
\setlength{\tabcolsep}{1.7mm}
\caption{Quantitative comparison of PSNR/SSIM on the Derf dataset for grayscale Gaussian video denoising, hearinafter, {\color{red} Red} and {\color{blue} Blue} indicate the best and the second best results, respectively.}
\label{tab:derf}
\begin{tabular}{ccccc}
\toprule
Derf & VBM4D\cite{maggioni2012video} & VNLB\cite{arias2018video} & VNLNet\cite{davy2021video} & FloRNN(Ours)\\
\midrule
$\sigma=10$ & 38.88/.9534 & \textcolor{blue}{40.57}/.9731 & 40.21/\textcolor{blue}{.9732} & \textcolor{red}{41.34}/\textcolor{red}{.9800}\\
$\sigma=20$ & 35.10/.9169 & \textcolor{blue}{36.81}/\textcolor{blue}{.9428} & 36.47/.9414 & \textcolor{red}{37.95}/\textcolor{red}{.9603}\\
$\sigma=40$ & 31.40/.8432 & \textcolor{blue}{32.95}/\textcolor{blue}{.8856} & 32.51/.8752 & \textcolor{red}{34.31}/\textcolor{red}{.9184}\\
Avg         & 35.13/.9045 & \textcolor{blue}{36.66}/\textcolor{blue}{.9338} & 36.40/.9299 & \textcolor{red}{37.87}/\textcolor{red}{.9529}\\
\bottomrule
\end{tabular}
\end{table}
\begin{table}[t]
\footnotesize
\setlength{\tabcolsep}{1.2mm}
\centering
\caption{Quantitative comparison of PSNR on the DAVIS dataset~\cite{pont20172017} for clipped Gaussian video denoising.}
\label{tab:davis}
\begin{tabular}{ccccc}
\toprule
DAVIS & ViDeNN~\cite{claus2019videnn} & FastDVDNet~\cite{tassano2020fastdvdnet} & PaCNet~\cite{vaksman2021patch} & FloRNN(Ours)\\
\midrule
$\sigma=10$ & 37.13 & 38.65 & \textcolor{red}{40.13} & \textcolor{red}{40.13}\\
$\sigma=30$ & 32.24 & 33.59 & \textcolor{blue}{34.92} & \textcolor{red}{35.81}\\
$\sigma=50$ & 29.77 & 31.28 & \textcolor{blue}{32.15} & \textcolor{red}{33.54}\\
Avg         & 33.05 & 34.51 & \textcolor{blue}{35.73} & \textcolor{red}{36.49}\\
\bottomrule
\end{tabular}
\end{table}
We also evaluate FloRNN on grayscale videos and on clipped Gaussian noise.
FloRNN shows compelling results in comparison to other methods. As shown in Table~\ref{tab:derf}, FloRNN outperforms  VNLNet~\cite{davy2021video} by 1.47dB in average on Derf\footnote{\url{https://media.xiph.org/video/derf}} dataset.
For clipped Gaussian noise, as shown in Table~\ref{tab:davis}, we achieve average PSNR of 0.76dB gain over PaCNet~\cite{vaksman2021patch} on DAVIS dataset~\cite{pont20172017}.
Figs.~\ref{fig:Set8}, \ref{fig:davis}, \ref{fig:crvd}, \ref{fig:iocv1}, \ref{fig:iocv2} show more qualitative results on Set8~\cite{tassano2019dvdnet}, DAVIS~\cite{pont20172017}, CRVD~\cite{yue2020supervised} and IOCV~\cite{kong2020comprehensive}, respectively.

\begin{figure*}
\centering
\includegraphics[width=\linewidth]{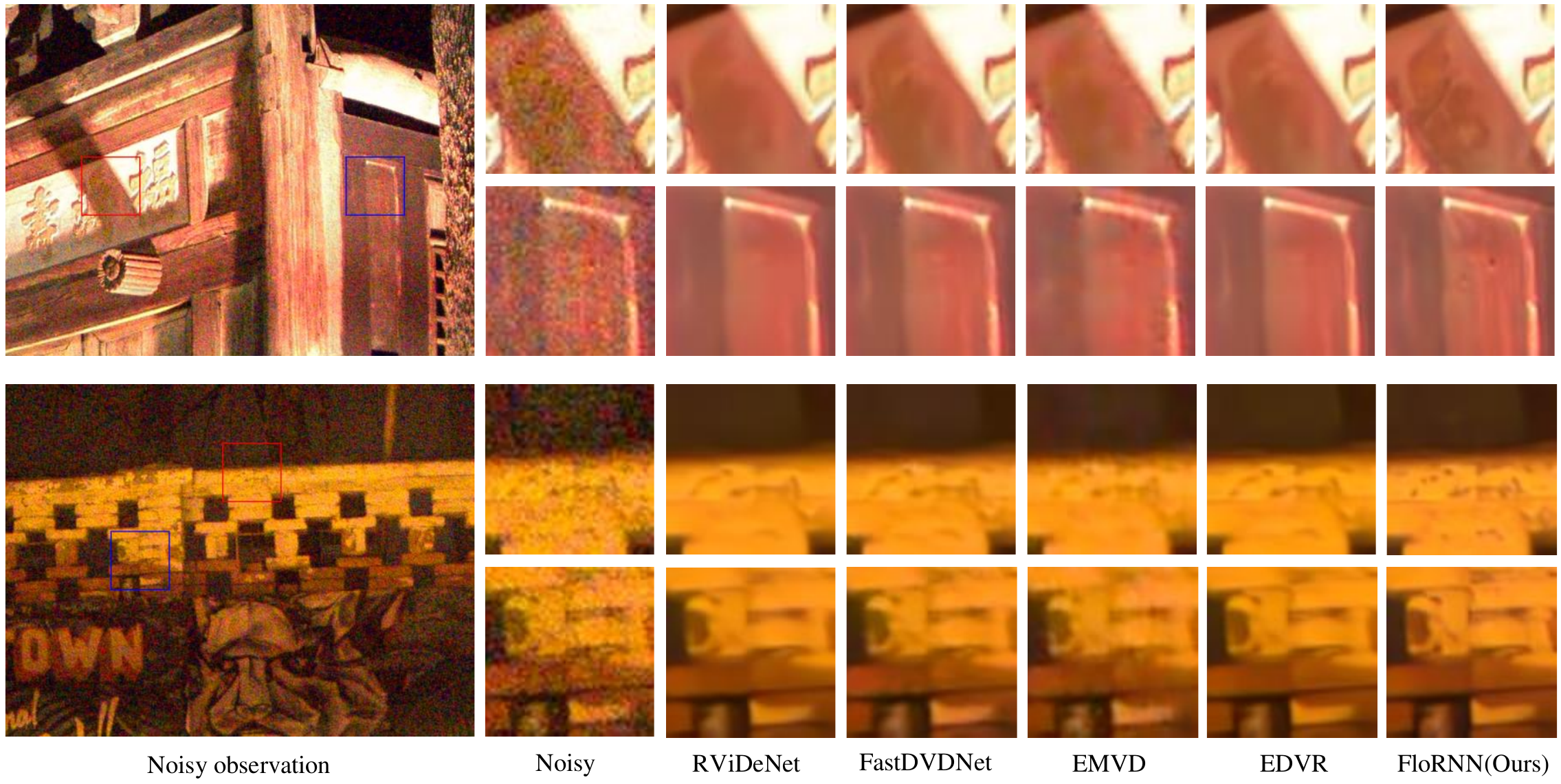}
\caption{More visual comparison of an outdoor scene on the CRVD dataset~\cite{yue2020supervised}.}
\label{fig:crvd}
\end{figure*}
\begin{figure*}[htbp]
\centering
\begin{subfigure}[b]{0.32\linewidth}
    \centering
    \includegraphics[width=\linewidth]{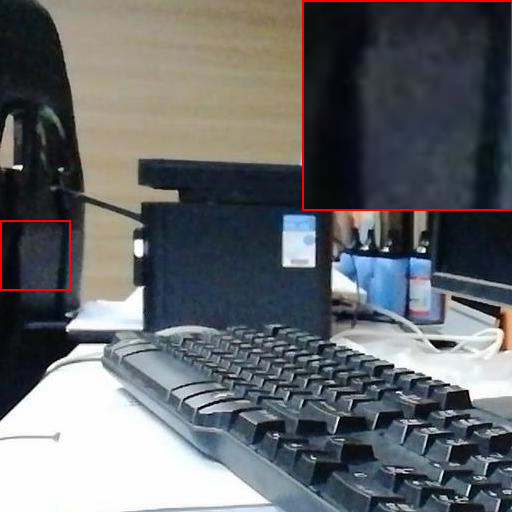}
    \caption{Noisy}
\end{subfigure}
\hfill
\begin{subfigure}[b]{0.32\linewidth}
    \centering
    \includegraphics[width=\linewidth]{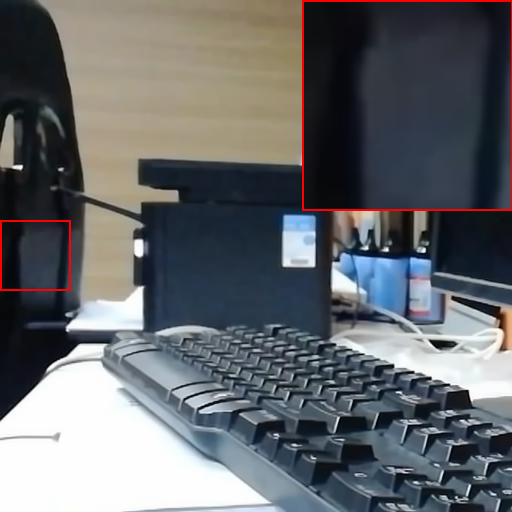}
    \caption{VBM4D}
\end{subfigure}
\hfill
\begin{subfigure}[b]{0.32\linewidth}
    \centering
    \includegraphics[width=\linewidth]{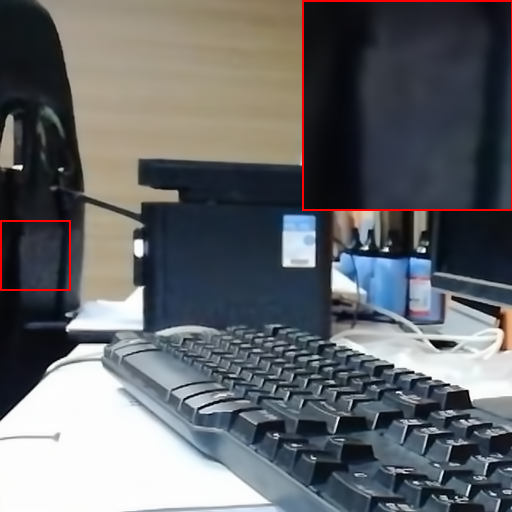}
    \caption{FastDVDNet}
\end{subfigure}
\\
\begin{subfigure}[b]{0.32\linewidth}
    \centering
    \includegraphics[width=\linewidth]{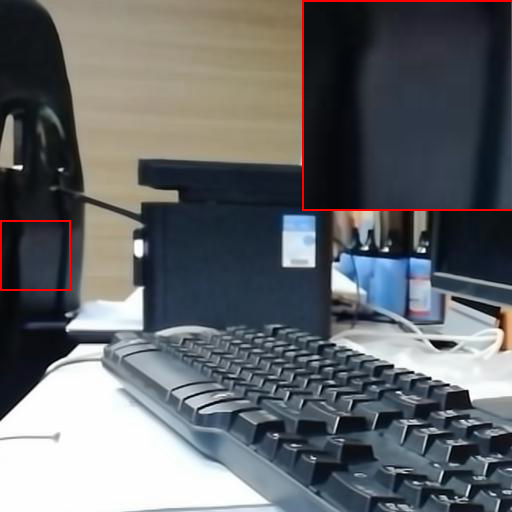}
    \caption{GT}
\end{subfigure}
\hfill
\begin{subfigure}[b]{0.32\linewidth}
    \centering
    \includegraphics[width=\linewidth]{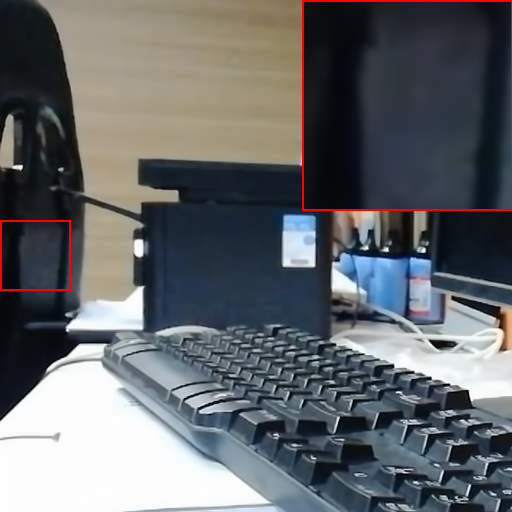}
    \caption{VNLNet}
\end{subfigure}
\hfill
\begin{subfigure}[b]{0.32\linewidth}
    \centering
    \includegraphics[width=\linewidth]{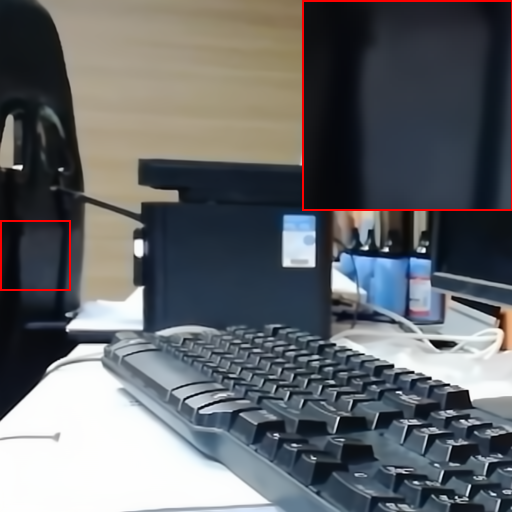}
    \caption{FloRNN(Ours)}
\end{subfigure}
\caption{More visual comparison on the IOCV dataset~\cite{kong2020comprehensive}.}
\label{fig:iocv1}
\end{figure*}
\begin{figure*}[htbp]
\centering
\begin{subfigure}[b]{0.32\linewidth}
    \centering
    \includegraphics[width=\linewidth]{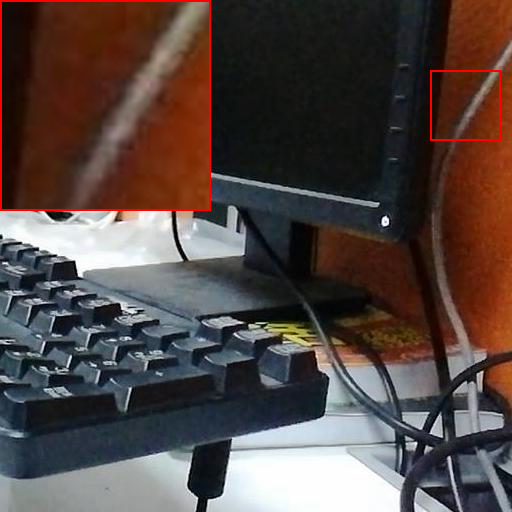}
    \caption{Noisy}
\end{subfigure}
\hfill
\begin{subfigure}[b]{0.32\linewidth}
    \centering
    \includegraphics[width=\linewidth]{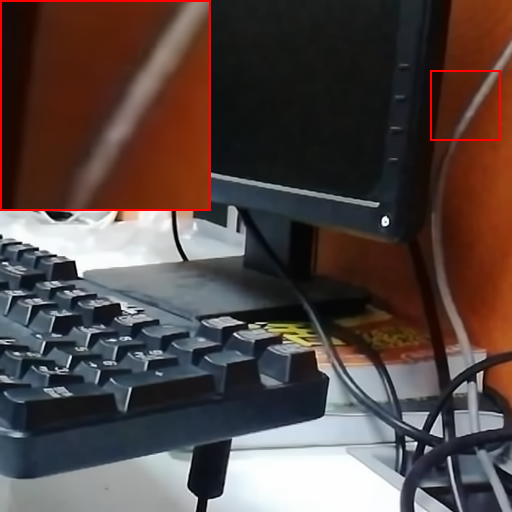}
    \caption{VBM4D}
\end{subfigure}
\hfill
\begin{subfigure}[b]{0.32\linewidth}
    \centering
    \includegraphics[width=\linewidth]{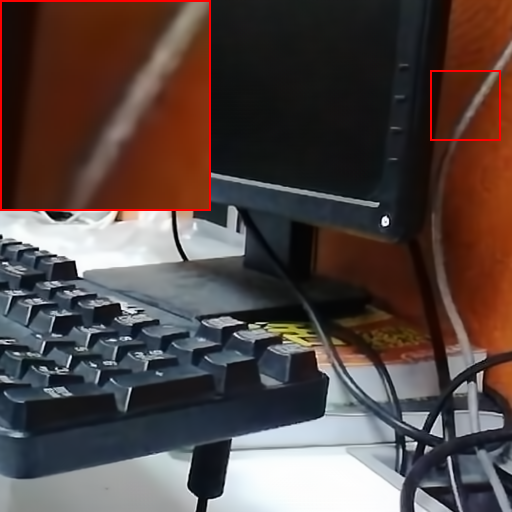}
    \caption{FastDVDNet}
\end{subfigure}
\\
\begin{subfigure}[b]{0.32\linewidth}
    \centering
    \includegraphics[width=\linewidth]{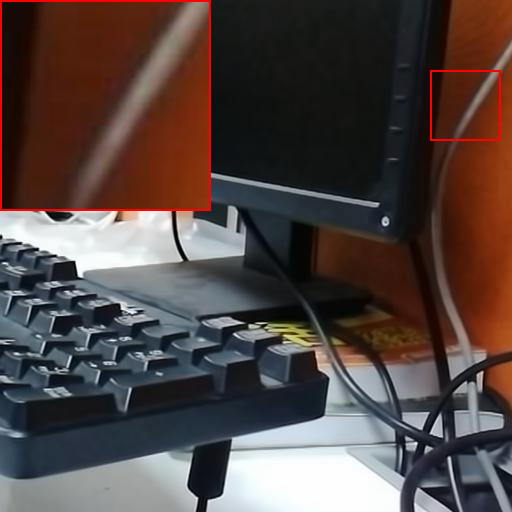}
    \caption{GT}
\end{subfigure}
\hfill
\begin{subfigure}[b]{0.32\linewidth}
    \centering
    \includegraphics[width=\linewidth]{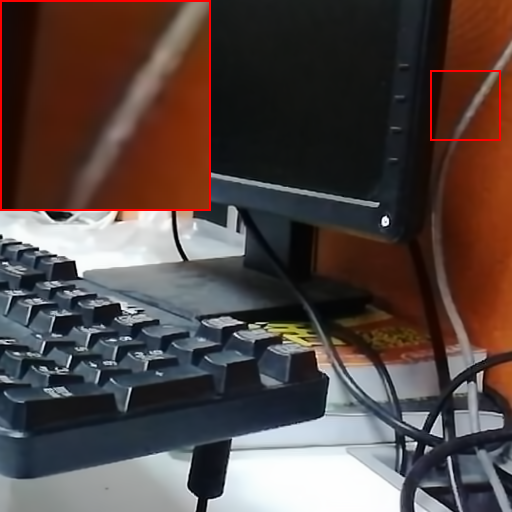}
    \caption{VNLNet}
\end{subfigure}
\hfill
\begin{subfigure}[b]{0.32\linewidth}
    \centering
    \includegraphics[width=\linewidth]{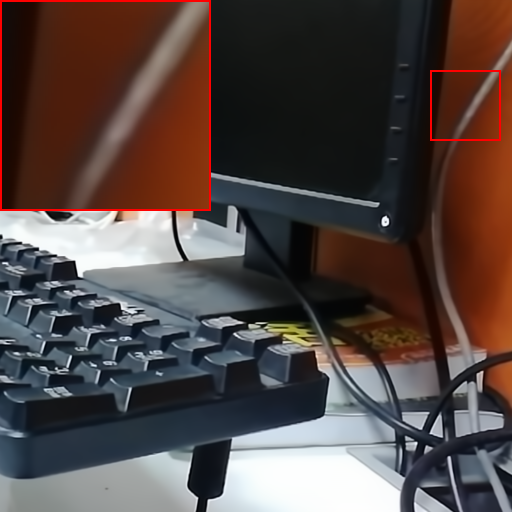}
    \caption{FloRNN(Ours)}
\end{subfigure}
\caption{More visual comparison on the IOCV dataset~\cite{kong2020comprehensive}.}
\label{fig:iocv2}
\end{figure*}

\end{document}